%% file: main.tex
\documentclass[preprint,5p,times,twocolumn]{elsarticle}

\usepackage{graphicx}
\usepackage{listings}
\usepackage[tight,footnotesize]{subfigure}
\usepackage{amsmath}
\usepackage{textcomp}

\usepackage[utf8]{inputenc}
\usepackage{tabularx}
\usepackage{xspace}
\usepackage{pifont}
\usepackage{epstopdf}
\usepackage{amssymb}
\usepackage{gensymb}
\usepackage{url}
\usepackage{multirow}
\usepackage{ctable}
\graphicspath{{./figures/}}
\usepackage{adjustbox,lipsum}
\usepackage[font=footnotesize]{subfig}
\usepackage{multirow}
\usepackage{amssymb}
\usepackage{epstopdf}
\usepackage{color,soul}

\newcommand{\ignore}[1]{}
\usepackage{algorithm,algpseudocode}
\usepackage{pdfpages}

\usepackage[nounderscore]{syntax}
\usepackage{etoolbox} 
\makeatletter

\newcommand{\newalgname}[1]{%
  \renewcommand{\ALG@name}{#1}%
}
\newalgname{Grammar}

\def\gr@implnumbereditem<#1> #2 {%
  \stepcounter{grammarline}%
  \sbox\z@{\hskip\labelsep\grammarlabel{#1}{#2}}
  \strut\@@par%
  \vskip-\parskip%
  \vskip-\baselineskip%
  \hrule\@height\z@\@depth\z@\relax%
  \item[%
    \rlap{\hskip\dimexpr\linewidth+\grammarindent\relax 
      \llap{(\uppercase\expandafter{\romannumeral 0\thegrammarline})}}%
    \unhbox\z@]%
  \catcode`\<\active%
}
\let\numberedgrammar\grammar

\pretocmd\numberedgrammar{\setcounter{grammarline}{0}}{}{}
\patchcmd\numberedgrammar
  {\gr@implitem}
  {\gr@implnumbereditem}
  {}{}
\patchcmd\numberedgrammar
  {\def\alt{\\\llap{\textbar\quad}}}
  {\let\alt\alt@num}
  {}{}

\newcounter{grammarline}
\makeatother

\journal{Applied Soft Computing}

\begin{document}

\begin{frontmatter}
\title{Runtime Data Center Temperature Prediction using Grammatical Evolution Techniques}

\author[ucm,lsi,ccs]{Marina Zapater\corref{cor1}}
\ead{marina.zapater@ucm.es}
\author[ucm]{José L. Risco-Martín}
\ead{jlrisco@ucm.es}
\author[lsi,ccs]{Patricia Arroba}
\ead{parroba@die.upm.es}
\author[ucm]{José L. Ayala}
\ead{jayala@ucm.es}
\author[lsi,ccs]{José M. Moya}
\ead{josem@die.upm.es}
\author[ucm]{Román Hermida}
\ead{rhermida@ucm.es}

\cortext[cor1]{Corresponding author}
\address[ucm]{DACYA, Universidad Complutense de Madrid, Madrid 28040, Spain}
\address[ccs]{CCS - Center for Computational Simulation, 
  Campus de Montegancedo UPM, 28660, Spain}
\address[lsi]{LSI - Integrated Systems Lab., Universidad Politécnica
  de Madrid, Madrid 28040, Spain}

\begin{abstract}
  Data Centers are huge power consumers, both because of the energy
  required for computation and the cooling needed to keep servers
  below thermal redlining. The most common technique to minimize
  cooling costs is increasing data room temperature. However, to avoid
  reliability issues, and to enhance energy efficiency, there is a
  need to predict the temperature attained by servers under variable
  cooling setups. Due to the complex thermal dynamics of data rooms,
  accurate runtime data center temperature prediction has remained as
  an important challenge.  By using Gramatical Evolution techniques,
  this paper presents a methodology for the generation of
  temperature models for data centers and the runtime prediction of
  CPU and inlet temperature under variable cooling setups. As opposed
  to time costly Computational Fluid Dynamics techniques, our models
  do not need specific knowledge about the problem, can be used in
  arbitrary data centers, re-trained if conditions change and have
  negligible overhead during runtime prediction.  Our models have
  been trained and tested by using traces from real Data Center
  scenarios. Our results show how we can fully predict the temperature
  of the servers in a data rooms, with prediction errors below
  2$\degree$C and 0.5$\degree$C in CPU and server inlet temperature
  respectively.
\end{abstract}


\end{frontmatter}


\section{Introduction}
\label{sec:intro}
\input{intro}

\section{Problem description}
\label{sec:problem}
\input{problem}

\section{Literature overview}
\label{sec:related}
\input{related}

\section{Gramatical evolution techniques}
\label{sec:modeling}
\input{modeling}

\section{Experimental Methodology}
\label{sec:method}
\input{method}

\section{Results}
\label{sec:results}
\input{results}

\section{Discussion}
\label{sec:discussion}
\input{discussion}

\section{Conclusions}
\label{sec:conclusions}
\input{conclusions}

\section*{Appendix}
\input{appendix}

\section*{Acknowledgments}
Research by Marina Zapater has been partly supported by a PICATA
predoctoral fellowship of the Moncloa Campus of International
Excellence (UCM-UPM). This project has been partially supported by
the Spanish Ministry of Economy and Competitiveness, under contracts
TEC2012-33892, IPT-2012-1041-430000 and RTC-2014-2717-3. The authors
thankfully acknowledge the computer resources, technical expertise and
assistance provided by the Centro de Supercomputación y Visualización
de Madrid (CeSViMa).

\section*{References}
\bibliographystyle{elsarticle-num}
\bibliography{main.bib}

\end{document}

%% file: intro.tex

Data Centers are found in every sector of the economy
and provide the computational infrastructure to support a wide range
of applications, from traditional applications to High-Performance
Computing or Cloud services. Over the past decade, both the
computational capacity of data centers and the number of these
facilities have increased tremendously without relative and
proportional energy efficiency, leading to unsustainable
costs~\cite{Kaplan:2008}. In 2010, data center electricity represented
1.3\% of all the electricity use in the world, and 2\% of all
electricity use in the US~\cite{thermal:koomey2011}. In year 2012,
global data center power consumption increased to 38GW, and in year
2013 there was a further rise of 17\% to 43GW~\cite{Venkatraman:2012}.

The cooling needed to keep the servers within reliable thermal
operating conditions is one of the major contributors to data center
power consumption, and accounts for over 30\% of the electricity
bill~\cite{Breen:ITHERM:2010} in traditional air-cooled
infrastructures. In the last years, both industry and academia have
devoted significant effort to decrease the cooling power, increasing
data center Power Usage Effectiveness (PUE), defined as the ratio
between total facility power and IT power. According to a report by
the Uptime Institute, average PUE improved from 2.5 in 2007 to 1.65 in
2013~\cite{uptimeInstitute:2013}, mainly due to more efficient cooling
systems and higher data room ambient temperatures.

However, increased room temperatures reduce the safety margins to CPU
thermal redlining and may cause potential reliability problems. To
avoid server shutdown, the maximum CPU temperature limits the minimum
cooling. The key question of how to set the supply temperature of the
cooling system to ensure the worst-case scenario, is still to be
clearly answered~\cite{El-Sayed:2012:TMD}. Most data centers typically
operate with server inlet temperatures ranging between 18$\degree$C
and 24$\degree$C, but we can find some of them as cold as 13$\degree$C
~\cite{Brandon:GoingGreen:07}, and others as hot as
35$\degree$C~\cite{Miller:Google:2012}. These values are often chosen
based on conservative suggestions provided by manufacturers, and
ensure inlet temperatures within the ranges published by ASHRAE (i.e.,
15$\degree$C to 32$\degree$C for enterprise
servers~\cite{ASHRAE:2011}).

Data center designers have collided with the lack of accurate models
for the energy-efficient real-time management of computing
facilities. One modeling barrier in these scenarios is the
high number of variables potentially correlated with temperature that
prevent the development of macroscopic analytical models. 
Nowadays, to simulate the inlet temperature of servers
under certain cooling conditions, designers rely on time consuming and
very expensive Computational Fluid Dynamics (CFD) simulations. These
techniques use numerical methods to solve the differential equations
that drive the thermal dynamics of the data room. They need to
consider a comprehensive number of parameters both from the server and
the data room (i.e. specific characteristics of servers such as
airflow rates, data room dimensions and setup). Moreover, they
are not robust to changes in the data center (i.e. rack placement and
layout changes, server turn-off, inclusion of new servers, etc.). If
the simulation fails to properly incorporate a relevant parameter, or
if there is a deviation between the theoretical and the real values,
the simulation becomes inaccurate. Due to the high economic and
computational cost of CFD simulation, models cannot be re-run each
time there is a change in the data room.

To minimize cooling costs, the development of models that accurately
predict the CPU temperature of the servers under variable
environmental conditions is a major challenge. These models need to
work on runtime, adapting to the changing conditions of the data room
automatically re-training if data center conditions change dramatically,
and enabling data center operators to increase room temperature
safely.

The nature of the problem suggests the usage of meta-heuristics
instead of analytical solutions. Meta-heuristics make few assumptions
about the problem, providing good solutions even when they have
fragmented information. Some meta-heuristics such as Genetic
Programming (GP) perform Feature Engineering (FE), a particularly
useful technique to select the set of features and combination of
variables that best describe a model. Grammatical Evolution (GE) is an
evolutionary computation technique based on GP used to perform
symbolic regression~\cite{Ryan:LNCS:1998}. This technique is
particularly useful to provide solutions that include non-linear terms
offering Feature Engineering capabilities and removing analytical
modeling barriers. Also, designer's expertise is not required to
process a high volume of data as GE is an automatic method. However,
GE provides a vast space of solutions that may need to be bounded to
achieve algorithm efficiency.


This paper develops a data center room thermal modeling methodology
based on GE to predict on runtime, and with sufficient anticipation,
the critical variables that drive reliability and cooling power
consumption in data centers. Particularly, the main contributions of
our work are the following:
\begin{itemize}
\item The development of multi-variable models that incorporate time
  dependence based on Grammatical Evolution to predict CPU and inlet
  temperature of the servers in a data room during runtime. Due to the
  feature engineering and symbolic regression performed by GE, our
  models incorporate the optimum selection of representative features
  that best describe the thermal behavior.
\item We prevent premature convergence by means of Social Disaster
  Techniques and Random Off-Spring Generation, dramatically reducing
  the number of generations needed to obtain accurate solutions. We
  tune the models by selecting the optimum parameters and fitness
  function using a reduced experimental setup, consisting of real
  measurements taken from a single server isolated in a fully
  sensorized data room.
\item We offer a comparison with other techniques commonly used in
  literature to solve temperature modeling problems, such as
  autoregressive moving average (ARMA) models, linear model
  identification methods (N4SID), and dynamic neural networks (NARX).
\item The proposal of an automatic data room thermal
  modeling methodology that scales our solution to a realistic Data
  Center scenario. As a case study, we model CPU and inlet
  temperatures using real traces from a production data center.

\end{itemize}
Our work allows the generation of accurate temperature
models able to work on runtime and adapt to the ever changing
conditions of these scenarios, while achieving very low average errors of
2$\degree$C for CPU temperature and 0.5$\degree$C for inlet temperature.

The remainder of the paper is organized as follows:
Section~\ref{sec:problem} accurately describes the
modeling problem, whereas Section~\ref{sec:related} provides an
overview of the current solutions. Section~\ref{sec:modeling}
describes our proposed solution, whereas Section~\ref{sec:method}
presents the experimental methodology. Section~\ref{sec:results} shows the
results obtained and Section~\ref{sec:discussion} discusses them. 
Finally, Section~\ref{sec:conclusions} concludes the paper.


%% file: problem.tex

\subsection{Data room thermal dynamics}
\begin{figure}
  \centering
  \includegraphics[width=1.0\columnwidth]{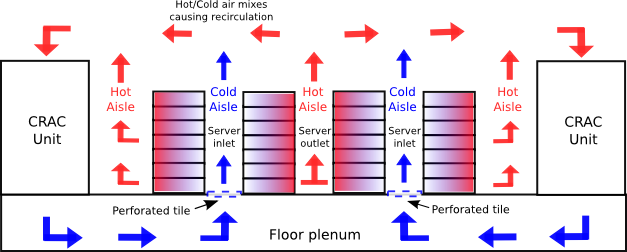}
  \caption{Typical raised-floor air-cooled Data Center layout}
  \label{fig:dc_cooling}
  \vspace{-0.10in}
\end{figure}

To ensure the safe operation of a traditional raised-floor air-cooled
Data Center, data rooms are equipped with
chilled-water Computer Room Air Conditioning (CRAC) units that use
conventional air-cooling methods. Servers are mounted in racks on a
raised floor. Racks are arranged in alternating cold/hot aisles, with
server inlets facing cold air and outlets creating hot
aisles. CRAC units supply air at a certain temperature and air
flow rate to the Data Center through the floor plenum. The floor has
some perforated tiles through which the blown air comes out. Cold air
refrigerates servers and heated exhaust air is returned to the CRAC
units via the ceiling, as shown in Figure~\ref{fig:dc_cooling}.


Even though this solution is very inefficient in terms of energy
consumption, the majority of the data centers use this
mechanism. In fact, despite the recent advances in high-density
cooling techniques, according to a survey by the Uptime Institute, in
2012 only 19\% of large scale data centers had incorporated other
cooling mechanisms~\cite{uptimeInstitute:2013}.
In some scenarios, the control knob of the cooling subsystem is the
cold air supply temperature, whereas in others, it is the return
temperature of the heated exhaust air to the CRAC unit.

The maximum IT power density that can be deployed in the Data Center
is limited by the perforated tile airflow. Because the plenum is
usually obstructed (e.g. blocked with cables in some areas), a
non-uniform airflow distribution is generated and each tile exhibits a
different pressure drop. Moreover, in data centers where the hot and
cold aisles are not isolated, which is the most common scenario, the
heated exhaust air recirculates to the cold aisle, mixing with the
cold air.

\subsection{Temperature-energy tradeoffs}

The factor limiting minimum data room cooling is maximum
server CPU temperature. Temperatures higher than 85$\degree$C can
cause permanent reliability
failures~\cite{thermal:AtienzaReliable2008}. At temperatures above
95$\degree$C, servers usually turn off to prevent thermal redlining.
Previous work on server power and thermal
modeling~\cite{Zapater:TPDS:2014}, shows how CPU temperature is
dominated by: i) power consumption, which is dependent on workload
execution, ii) fan speed, which changes the cooling capacity of the
server, and iii) server cold air supply (inlet temperature).

Thus, to keep all the equipment under normal operation, CRAC units
have to supply the air at an adequately low temperature to ensure that
all CPU's are below the critical threshold.
However, inlet temperature is also not uniform across servers. The
cold air temperature at the server inlet depends on several
parameters: i) the CRAC cold air supply, ii) the airflow
rate through the perforated tiles, and iii) the outlet temperature of
adjacent servers due to air recirculation.

Setting the cooling air supply temperature to a low value, even
though ensures safety operation, implies increased power consumption
due to a larger burden on the chiller system. The goal of
energy-efficient cooling strategies is to increase the cold air supply
temperature without reaching thermal redlining. Due to the non-linear
efficiency of cooling systems, lowering air supply temperature can
yield important energy savings. A metric widely used is that each
degree of increase in air supply temperature yields 4\% energy savings
in the cooling subsystem~\cite{Miller:2007}. To increase air supply
temperature safely, however, we need to predict not only the inlet
temperature to the servers, but also the CPU temperature that each
server attains under the current workload.

Due to the temperature gradients between hot and cold aisles and the
data room layout and geometry, the air inside a data center behaves
like a turbulent fluid. Thus, obtaining an analytical relation
between cold air supply and server inlet temperature is not trivial,
making inlet and CPU temperature prediction a challenging
problem. Besides, data centers are composed of thousands of CPU cores,
whose temperatures need to be modeled independently. This prevents
the usage of classical regression techniques that need human
interaction to train and validate the models.


%% file: related.tex

Data center room thermal modeling enables both thermal emergency
management and energy optimization, and enhances reliability. Because
of the turbulent behavior of the air in the Data Center room,
Computational Fluid Dynamics (CFD) simulation has traditionally been
the most commonly used solution in both industry and
academia~\cite{Marshall:WP:2011}.

CFD is used to model the inlet and outlet temperature of servers,
given cold air supply parameters, room layout, server configuration
and utilization, in order to either optimize cooling costs or detect
hot spots~\cite{Abbasi:HPDC:2010}.
CFD solvers perform a three-dimensional numerical analysis of the
thermodynamic equations that govern the data room. Their main
drawbacks is that they require and expert to configure the
simulation, and are computationally costly both in the
modeling stage (i.e. modeling a small-sized data room may take from
hours to days) and in the evaluation phase, thus preventing their online
usage. Moreover, CFD simulation is not robust to changes in the layout
of the Data Center, i.e. server placement, open tiles,
workload running or cold air supply setting.

To solve these issues,\ignore{recent research by Phan
  et.al.~\cite{Phan:JEB:2014} proposes the usage of Building Energy
  Simulation (BES) Programs to obtain an insight of data room behavior
  with less accuracy and computational costs than CFD.} Chen
et.al.~\cite{Chen:2012:RTSS} use CFD together with sensor information
to calibrate the simulation and reduce computational complexity. Their
work achieves a prediction error below 2$\degree$C when predicting
temperature 10 minutes in advance. Other
work~\cite{Abbasi:Springer:2013} presents the Data Center as a
distributed Cyber-Physical System (CPS) in which both computational
and physical parameters can be measured with the goal of minimizing
energy consumption. Our work leverages this concept by using a
monitoring system~\cite{Pagan:DCIS:2013} capable of collecting 
environmental (i.e. cold air supply and server inlet 
temperature, airflow, etc.) and server data (i.e. temperature,
power, fan speed, etc.) from a real data center.

A common alternative to CFD are abstract heat flow models. These
models characterize the steady state of hot air recirculation
inside the data center. Recirculation is described by a
cross-interference coefficient matrix which denotes how much of every
node's outlet heat contributes to the inlet of every other node. This
matrix is obtained in an offline profiling stage using
CFD~\cite{Varsamopoulos:Springer:2009}. Even though profiling is still
costly, model evaluation can be performed online.

Machine learning techniques have also been used in Data Center
modeling. The Weatherman~\cite{Moore:ICAC:2006} tool uses neural
networks to predict the inlet temperature of servers, obtaining
prediction errors below 1$\degree$C in over 90\% of their
traces. However, they use simulation traces obtained with CFD
simulation for their training and test sets, instead of real data. The
problem behind time series prediction can be explained as a problem of
extracting a manageable set of adequate features, followed by a
regression mechanism. Careful selection of features and their horizon
is therefore of much greater importance compared with the static-data
prediction problem. Neural network-based approaches require previous
knowledge of the parameters that drive thermal modeling, obtaining
them using pseudo-exhaustive algorithms. Our work, on the contrary,
relies on the benefits of feature engineering in Symbolic Regression
problems to obtain the relevant features and construct the models in
an automated way.

The work by De Silva \emph{et al.}~\cite{Silva:2013:ICMLA} is the one
most similar to ours, regarding the modeling methodology. The authors
use Grammatical Evolution for electricity load prediction. As opposed
to our work, this paper is focused on predicting the trend, momentum
and volatility indicators of a timeseries, not on obtaining a physical
model, i.e. they do not solve a multi-variable problem.

In summary, the main issues in all previous approaches are: i) they
monitor and predict inlet temperature instead of CPU temperature, ii)
modeling is performed for only certain hand-picked cooling and
workload configurations, iii) the use of CPU utilization as a proxy
for server power, iv) they assume data centers with homogeneous
servers, v) server fan speed is considered constant, vi) results
are not validated with real traces, and vii) model construction
requires specific knowledge on the problem and classical feature
selection, which prevents the usage of automated techniques.

Our work, on the contrary, first predicts inlet temperature and then
uses this result to predict CPU temperature, which is the factor
limiting cooling. Both in our training and test sets, we use real
traces obtained from enterprise servers in a data center. Moreover, as
shown in previous work~\cite{Zapater:TPDS:2014}, in highly
multi-threaded enterprise servers, utilization is not a good proxy for
power for arbitrary workloads. 

Enterprise servers come with automatic temperature-driven variable fan
control policies. When fan speed changes, so does the airflow and the
server cooling capacity~\cite{Patterson:ITHERM:2008}. Our methodology
also considers the contribution of variable fan speed, allowing us to
predict temperature in heterogeneous data center setups running
arbitrary workloads.

At the server level, Heather et al.~\cite{Heath:ASPLOS:2006} propose
a server temperature prediction model based on simplified
thermodynamic equations obtaining results within 1$\degree$C of
accuracy. Even though this approach predicts CPU
temperature and takes into account inlet temperature, it does not
predict the inlet and needs specific knowledge about several server
parameters. Our approach only uses data from the generic sensors
deployed in the server and data room.

Another common approach to CPU temperature modeling is the usage of
autoregressive moving average (ARMA) modeling to estimate future
temperature accurately based on previous
measurements~\cite{Coskun:TCAD:2009}. Their main drawbacks are that,
because they only use past temperature samples, the prediction horizon
is usually below one second. Moreover, they do not provide a physical
model, disregarding the effect of power or airflow, and need to be
retrained often.

As opposed to others, our work achieves prediction horizons of 1 minute for
CPU temperature and 10 minutes for inlet temperature with high
accuracy. This enables data center operations to take action before
thermal events occur, by changing either the workload or the cooling
in the data center. 


%% file: modeling.tex
Evolutionary algorithms use the principles of evolution to turn one
population of solutions into another, by means of selection, crossover
and mutation. Among them, Genetic Programming (GP) has proven to be
effective in a number of Symbolic Regression (SR)
problems~\cite{Vladislavleva:TEVC:2009}. However, GP presents some
limitations like bloating of the evolution (excessive growth of memory
computer structures), often produced in the phenotype of the
individual. In the last years, variants to GP like Grammatical
Evolution (GE) appeared as a simpler optimization
process~\cite{ONeill:TEVC:2001}. GE is inspired in the biological
process of generating a protein given the DNA of an
organism. GE evolves computer programs given a set of rules, adopting
a bio-inspired genotype-phenotype mapping process.

In this section, we describe how we perform feature selection, provide
a brief insight on the grammars and mapping process, as well as on
several model parameters.

\vspace{-0.1cm}
\subsection{Feature selection and model definition}

In this work we use Feature Engineering (FE) and Grammatical Evolution
to obtain a mathematical expression that models CPU and server inlet
temperature. This expression is derived from experimental measurements
in real server and data room scenarios, gathering data that have an
impact on temperature, according to previous work in the
area~\cite{Zapater:TPDS:2014,Pagan:DCIS:2013}. To predict CPU
temperature, we gather server power, fan speed, inlet
temperature and previous CPU temperature measurements. For inlet
temperature, we gather the CRAC air supply and return temperature,
humidity, pressure difference through perforated tiles (which is a
measurement that provides information about airflow) and previous inlet
temperature measurements. Our goal is to predict temperature a certain
time (samples) in advance, by using past data of the available
magnitudes within a window. We may use past samples from the magnitude
we need to predict, or even previously predicted data.

\begin{figure}
  \centering
  \includegraphics[width=0.9\columnwidth]{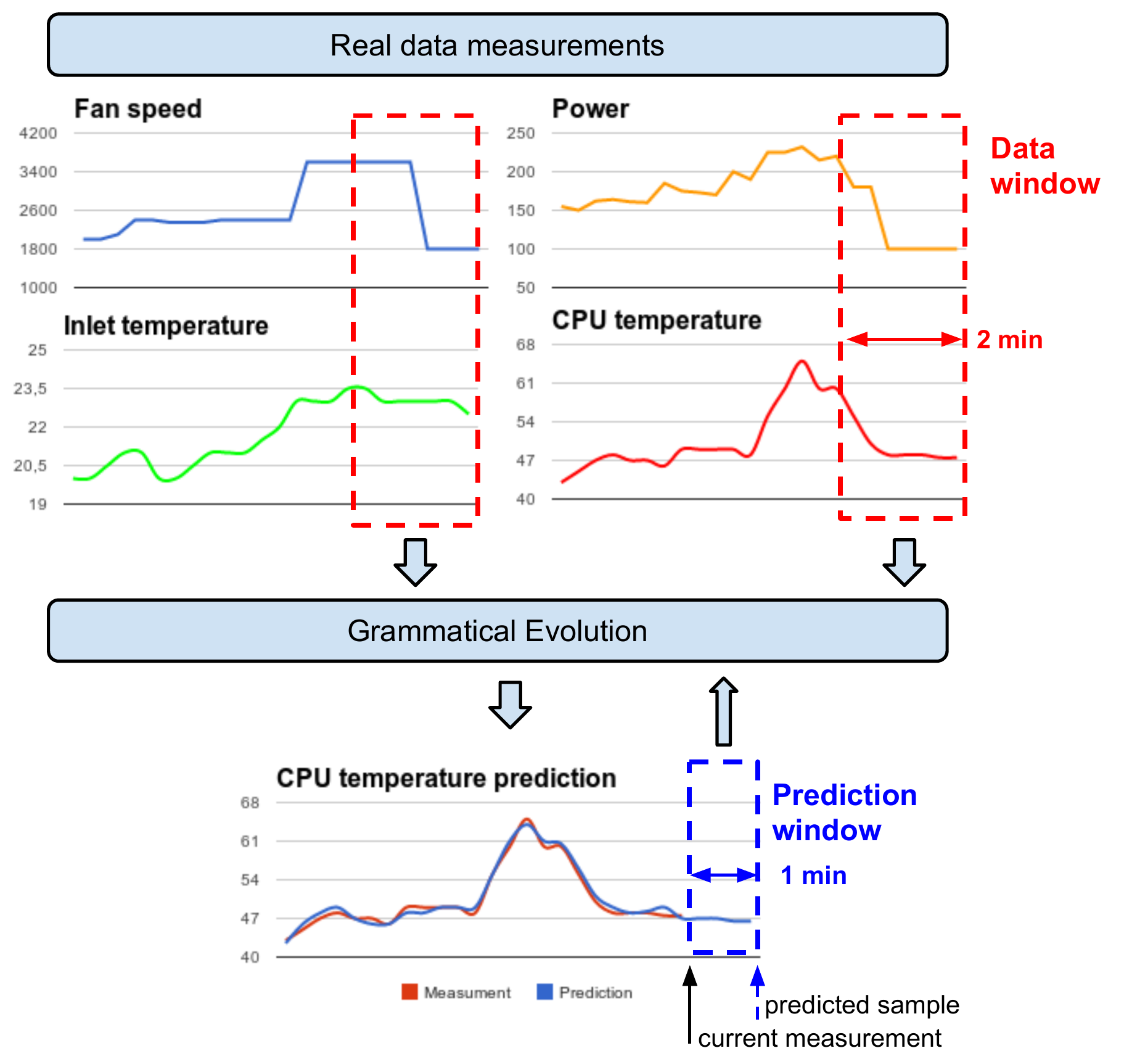}
  \caption{CPU temperature prediction diagram.\ignore{CPU temperature
  is predicted given past data measurements of various magnitudes (data
  window) and past CPU temperature predictions.}}
  \label{fig:prediction-diagram}
  \vspace{-0.2cm}
\end{figure}

For illustration purposes, in Figure~\ref{fig:prediction-diagram} we
show a diagram in which CPU temperature is predicted 1 minute ahead
given: i) 2 minutes of past measurements (data window) for fan speed,
server power, inlet and CPU temperature and, ii) the previous CPU
temperature predictions (prediction window).

Formally, we claim that CPU temperature prediction for a certain
time instant $\alpha$ samples into the future is a function of past
data measurements within a window of size
$i=\{0..W_{cpu}\}$, and previously predicted values within a window of
size $j=\{1..\alpha\}$ as expressed in Eq.(\ref{eq:cpu-temp}):

\vspace{-0.2cm}
\begin{equation}
\begin{split}
  \widehat{T}_{CPU}(k+\alpha) = & f\big( T_{inlet}(k-i), FS(k-i), P(k-i),\\
  & T_{CPU}(k-i), \widehat{T}_{CPU}(k+j)\big)
\end{split}
\label{eq:cpu-temp}
\end{equation}

\noindent where $T_{inlet}$ is a short form for the previous inlet
temperature values in a window: $\{T_{inlet}(k-i) | i \in \{0, ...,
    W_{cpu}\}\}$. $T_{CPU}$, $FS$ and $P$ are past CPU temperature, fan
speed, and server power consumption values respectively, which are
defined similarly, and $\widehat{T}_{CPU}$ are previous temperature
predictions.

For inlet temperature, our claim is that inlet temperature $T_{inlet}$
of a certain server is driven by the room thermal dynamics and can be
expressed as a function of the cold air supply (or return)
temperature, $T_{CRAC}$, differential pressure across perforated
tiles $\gamma$ (measured in inches of water, $in H_2O$) and data room
humidity $h$ (in percentage), as in Eq.(\ref{eq:inlet-temp}):

\begin{equation}
\begin{split}
  \widehat{T}_{inlet}(k+\beta) = & f\big( T_{CRAC}(k-i), 
  \gamma(k-i), h(k-i),\\
  & FS_{p-m}(k-i), \widehat{T}_{inlet}(k+j)\big)
\end{split}
\label{eq:inlet-temp}
\end{equation}

\noindent where the data window can be defined in the range
$i=\{0..W_{inlet}\}$ and the prediction window is $j=\{1..\beta\}$

Note that, in general, $\alpha$ and $\beta$ are not equal, as the room
dynamics are much slower than the CPU temperature dynamics of the
servers, i.e. in a real data room we might need hours to appreciate
substantial differences in ambient temperature, whereas CPU
temperature changes within seconds. The selection of relevant
features among all data measurements is a Symbolic Regression (SR)
problem.\ignore{ SR tries to simultaneously obtain a mathematical
  expression while including the relevant features to reproduce a set
  of discrete data.} In our approach, GE allows the generation of
mathematical models applying SR.

Regarding both the structure and the internal operators, GE works like
a classic Genetic Algorithm~\cite{Back:TEVC:1997}. GE evolves a
population formed by a set of individuals, each one constituted by a
chromosome and a fitness value. In SR, the fitness value is usually a
regression metric like Mean Squared Error (MSE). In GE, a chromosome
is a string of integers. In the optimization process, GA operators,
are iteratively applied to improve the fitness value of each
individual. In order to compute the fitness function for every
iteration and extract the mathematical expression given by an
individual (phenotype), a mapping process is applied to the chromosome
(genotype). This mapping process is achieved by defining a set of
rules to obtain the mathematical expression, using grammars in Backus
Naur Form (BNF)~\cite{ONeill:TEVC:2001}. 

The process does not only perform parameter identification. In
conjunction with a well-defined fitness function, the evolutionary
algorithm is also computing mathematical expressions with the set of
features that best fit the target system. Thus, GE is also defining
the optimal set of features that derive into the most accurate power
model.

Moreover, this methodology can be used to predict magnitudes with
memory, such as temperature, where the current observation depends on
past values. To incorporate time dependence, data used for model
creation needs to be a timeseries. In addition, we need to tune our
grammars so that they can produce models where past temperature values
can be used to predict temperature a certain number of samples into
the future.
Grammar~\ref{fig:bnf-time} shows an example where variable x may take
values in the current time step $k$, i.e., $x[k-0]$ or in previous samples
like $x[k-1]$ or $x[k-2]$. Moreover, a new variable $xpred[k-idx]$ can be
included, that accounts for previously predicted values of variable
$x$.

Including time dependence into a grammar has some drawbacks. First, we
are substantially increasing the search space of our algorithms, as
now the GE needs to search for the best solution among all variables
within the specified window. As a consequence, the number of
generations needed to obtain a good fitness value increases. Second,
as we show in the results section, depending on the prediction horizon
the models tend to fall into a local optimum, in which the best
phenotype is the last available observation of the predicted
variable. To address the latter challenge, we propose the use of the
premature convergence prevention techniques that are next explained,
and that also benefit the converge time of our algorithms. Despite the
drawbacks, introducing time dependence in our modeling is a must, as
temperature transients (both at the server and the data center level)
are not negligible and need to be accurately predicted.

For a more detailed explanation on the principles of the mapping
process, and how the BNF grammars are used to incorporate time
dependence, the reader is referred to the Appendix.

\begin{algorithm}
\begin{numberedgrammar}
<expr> ::= <expr><op><expr> | <preop>(<expr>) | <var> 

<op>   ::= +|-|*|/

<preop> ::= sin| cos | log

<var>  ::= x[k-<idx>] | xpred[k-<idx>]
               | y | z | <num>

<num>  ::= <dig>.<dig> | <dig>

<dig>  ::= 0 | 1 | 2 | 3 | 4 | 5

<idx>  ::= 0 | 1 | 2
\end{numberedgrammar}
\caption{Example of a grammar in BNF that generates phenotypes
  with time dependence}
\label{fig:bnf-time}
\end{algorithm}

\subsection{Preventing premature convergence}
Premature convergence of a genetic algorithm arises when the
chromosomes of some high rated individuals quickly dominate the
population, reducing diversity, and constraining it to converge to a
local optimum. Premature convergence is one of the major shortcomings
when trying to model low variability magnitudes by using GE
techniques.

To overcome the lack of variety in the population, work by Kureichick
et al.~\cite{Kureichick:1996} proposes the usage of Social Disaster
Techniques (SDT). This technique is based on monitoring the population
to find local optima, and apply an operator:
\begin{enumerate}
\item \emph{Packing}: all individuals having the same fitness
  value except one are fully randomized.
\item \emph{Judgment day}: only the fittest individual
  survives while the remaining are fully randomized.
\end{enumerate}

Work by Rocha et al.~\cite{Rocha:Springer:1999} proposes the usage of
Random Off-spring Generation (ROG) to prevent the crossover of two
individuals with equal genotype, as this would result in the
off-spring being equal to the parents. Individuals are
tested before crossover and, if equal, then one off-spring (1-RO) or
both of them (2-RO) are randomly generated.

Both previous solutions have shown important benefits in classical
Genetic Algorithms problems. In our work, we use these techniques to
improve the convergence time of our solutions, as we show in
Section~\ref{sec:results}.

\subsection{Fitness}
The goal of using GE for data room thermal modeling is to obtain
accurate models. Thus, our fitness function
needs to express the error resulting in the estimation process. To
measure the accuracy in our prediction, we would preferably use the
Mean Absolute Error (MAE). 
However, because temperature is a magnitude
that varies slowly and might remain constant during large time
intervals, we need to give higher weight to large errors. To this end,
the fitness function $f$ presented in Eq.(\ref{eq:fitness}) tries
to reduce the variance of the model, leading the evolution to obtain
solutions that minimize the the Root Mean Square Error (RMSE):

\begin{equation}
\label{eq:fitness}
f = \sqrt{\frac{1}{N} \cdot \sum_{i} {e_\mathrm{i}}^2}
\end{equation}

\noindent where the estimation error $e_\mathrm{i}$ represents the
deviation between the real temperature samples (both for CPU
and inlet temperature modeling) obtained by the monitoring
system $T$, and the estimation obtained by the model
$\widehat{T}$. $i$ represents each sample of the entire set of $N$
samples used to train the algorithms.

\subsection{Problem constraints}

As we are modeling the behavior of physical magnitudes for
optimization purposes, we need to obtain a solution with
physical meaning. To this end, we constrain the general problem of
temperature modeling in several ways that are subsequently presented,
while still being able to address heterogeneous workloads,
architectures and topologies. In the results section
we evaluate the impact of these constraints on the model generation
stage.

\subsubsection{Constraining the grammar} 
The mathematical expressions can be constrained to a limited
number of functions with physical meaning. Because
temperature exhibits exponential transients, we can include the
exponential function in our grammar, whereas we do not find physical
basis to include other mathematical functions such as sines or
cosines.

\subsubsection{Fitness biasing} 
Some parameters drive the variables being modeled. For instance, power
consumption drives CPU temperature. As we want to obtain models able
to capture the physical phenomena that drive temperature, this
magnitude should be present in the final model. Thus, CPU temperature
models that do not include power in their phenotype are expected to
provide good results in the training phase, but to perform poorly for
the test, as they are not capturing the physical phenomena. To solve
this issue, we can force the appearance of some parameters by biasing
the fitness, giving higher weights (i.e. worse fitness) to expressions
that miss a parameter. By biasing fitness we speed-up convergence, we
ensure that our models incorporate all parameters directly correlated
with temperature, but we could obtain less accurate results.

\subsubsection{Real vs. mixed models}
Purely real models only use real temperature data measurements to
predict future samples. Purely predictive models do not used previous
temperature measurements, but may use previous predictions. Mixed
models may used both real and predicted data. Adding the predicted
samples as a variable increases the size of the search
space but may provide higher accuracy.


%% file: method.tex
In this section we describe the experimental methodology followed in
this paper to model server and environmental parameters in Data
Centers.  First, we describe an scenario consisting only in the
temperature prediction of one server in a small air-cooled data
room. We use this scenario to tune the model parameters, testing those
that generate better models and studying the convergence of the
solutions. Then, we apply the best algorithm configuration to a
real data center. As a case study, we use real traces of CeSViMa
Data Center, a High Performance Computing cluster at Universidad
Politécnica de Madrid in Spain.

\subsection{Reduced scenario}
This scenario consists on an Intel Xeon RX-300 S6 server equipped with
1 quad-core CPU and 16GB of RAM. The server is installed in a rack
with another 4 servers, 2 switches and 2 UPS units, in an air-cooled
data room of approximately 30$m^2$, with the rack inlet facing the
cold air supply and the outlet to the heat exhaust. The cooling
infrastructure consists on a Daikin FTXS30 split that pumps cold air
from the ceiling, and there is no floor plenum. The cold air supply
ranges from 16$\degree$C to 26$\degree$C. The data room is
fully monitored, and both the cooling and server workload
are controllable.

\subsubsection{Monitoring}
Both the server and data room are fully monitored using the
internal server sensors and a wireless sensor network, as described
in~\cite{Pagan:DCIS:2013}. In particular, server CPU temperature and
fan speed values are obtained via the server internal sensors,
collected through the Intelligent Platform Management Interface (IPMI)
tool~\footnote{http://ipmitool.sourceforge.net/}. IPMI allows polling
the internal sensors of enterprise servers with negligible
overhead. As the server is not shipped with power consumption
sensors, we use non-intrusive current clamps connected to the power
cord of the server to gather total server power consumption. Wireless
sensors monitor the inlet temperature of the server, the cold air supply
temperature of the split unit and data room humidity. Data are sent to
the monitoring server, stored and aligned to ensure a common
timestamp.

\subsubsection{Training and test set generation}
We generate the training and test set by assigning a wide variety of
workloads that exhibit various stress levels in the CPU and
memory subsystems of the server while we modify the cold air supply
temperature of the split in a range from 16$\degree$C to
26$\degree$C. All workloads used are a representative set, in terms of
stress to the server subsystem and power consumption, of the ones
that can be found in High-Performance Computing data centers. Also,
the temperatures selected for cold air supply temperature are within the
allowable ranges in current data centers.

The workloads used are the following: i)
\emph{Lookbusy}\footnote{http://www.devin.com/lookbusy/}, a synthetic
workload that stresses the CPU to a customizable utilization value,
avoiding the stress of memory and disk; ii) a modified version of the
synthetic benchmark
\emph{RandMem}\footnote{http://www.roylongbottom.org.uk}, that allows
us to stress random memory regions of a given size with a given access
pattern, and iii) HPC workloads belonging to the SPEC CPU 2006
benchmark suite~\cite{SPECCPU}.

During training, we launch \emph{Lookbusy} and \emph{Randmem} at
various utilization values, plus a subset of the SPEC CPU 2006
benchmarks that exhibit a distinctive set of characteristics according
to Phansalkar et al.~\cite{Phansalkar:2007:SIGARCH}. Both the arrival
time and task duration are randomly selected. During execution, the
cold air supply temperature is also randomly changed.
For the test set, we randomly launch a SPEC CPU benchmark, with 
random waiting intervals while changing cold air supply
temperature.

\begin{figure}
  \centering
  \includegraphics[width=0.9\columnwidth]{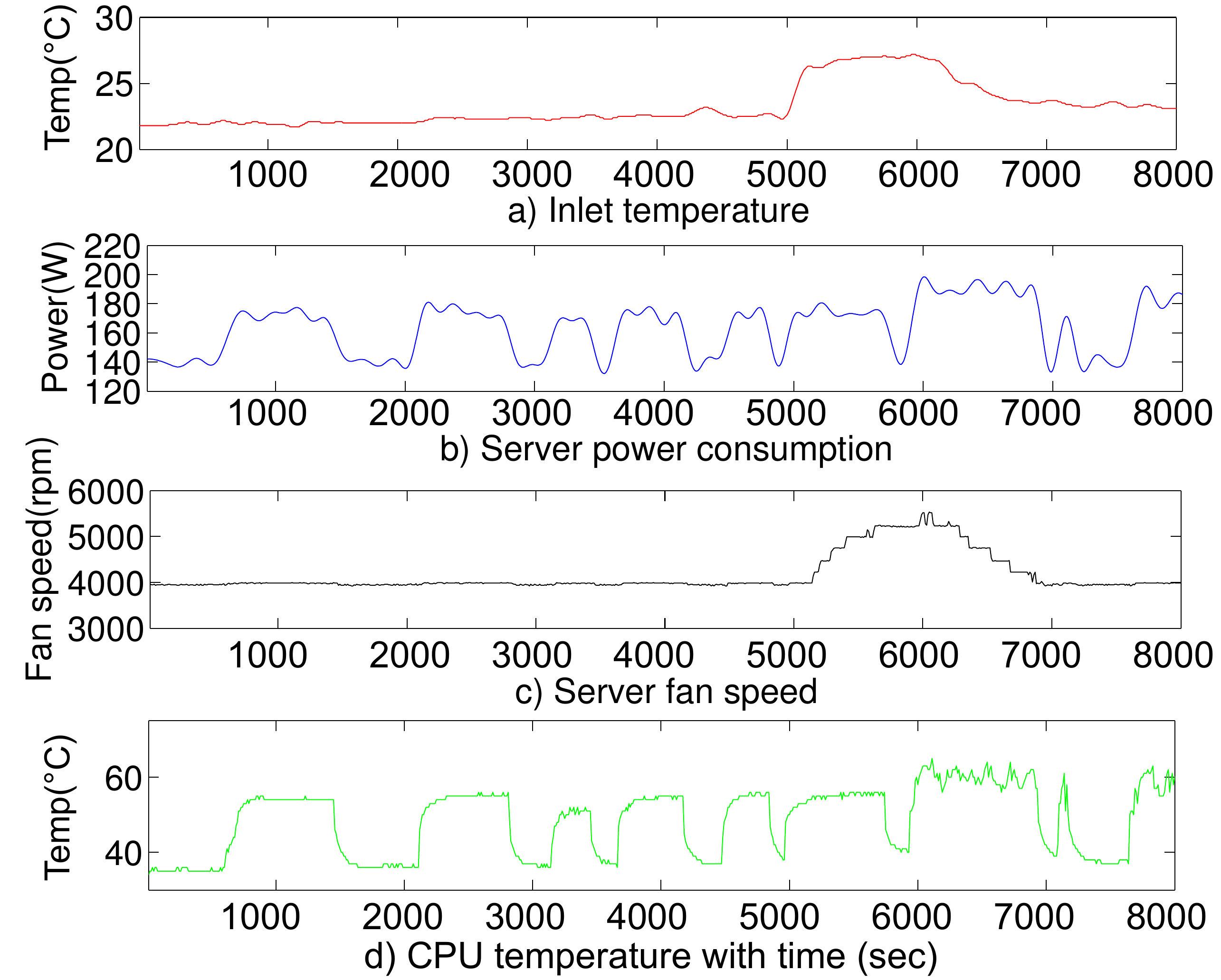}
  \caption{Training samples used for CPU temperature modeling}
  \label{fig:training-samples}
  \vspace{-0.1cm}
\end{figure}

Our monitoring system collects all data with a 10 second sampling
interval for a total time of 5 hours for the training and 10 hours for
the test set. Figure~\ref{fig:training-samples} shows part of the
training set used for modeling.

\subsection{Case study: CeSViMa Data Center}

To show how our solution can be applied to a real data center
scenario, this paper presents a case study for a real High-Performance
Computing Data Center at the Madrid Supercomputing and Visualization
Center (CeSViMa)\footnote{http://www.cesvima.upm.es/}. CeSViMa hosts
the \emph{Magerit} Supercomputer, a cluster consisting of 286 computer
nodes in 11 racks, providing 4,160 processors to execute
High-Performance jobs on demand. 245 of the 286 nodes are IBM PS702 2S
with 2 Power7 CPU's blade servers, each with 8 cores running at 3.3GHz
and 32GB of RAM. The other 41 nodes are HS22 2U servers with 2 Intel
Xeon processors of 8 cores each at 2.5GHz and 64GB RAM.

\begin{figure}
  \centering
  \includegraphics[width=1.0\columnwidth]{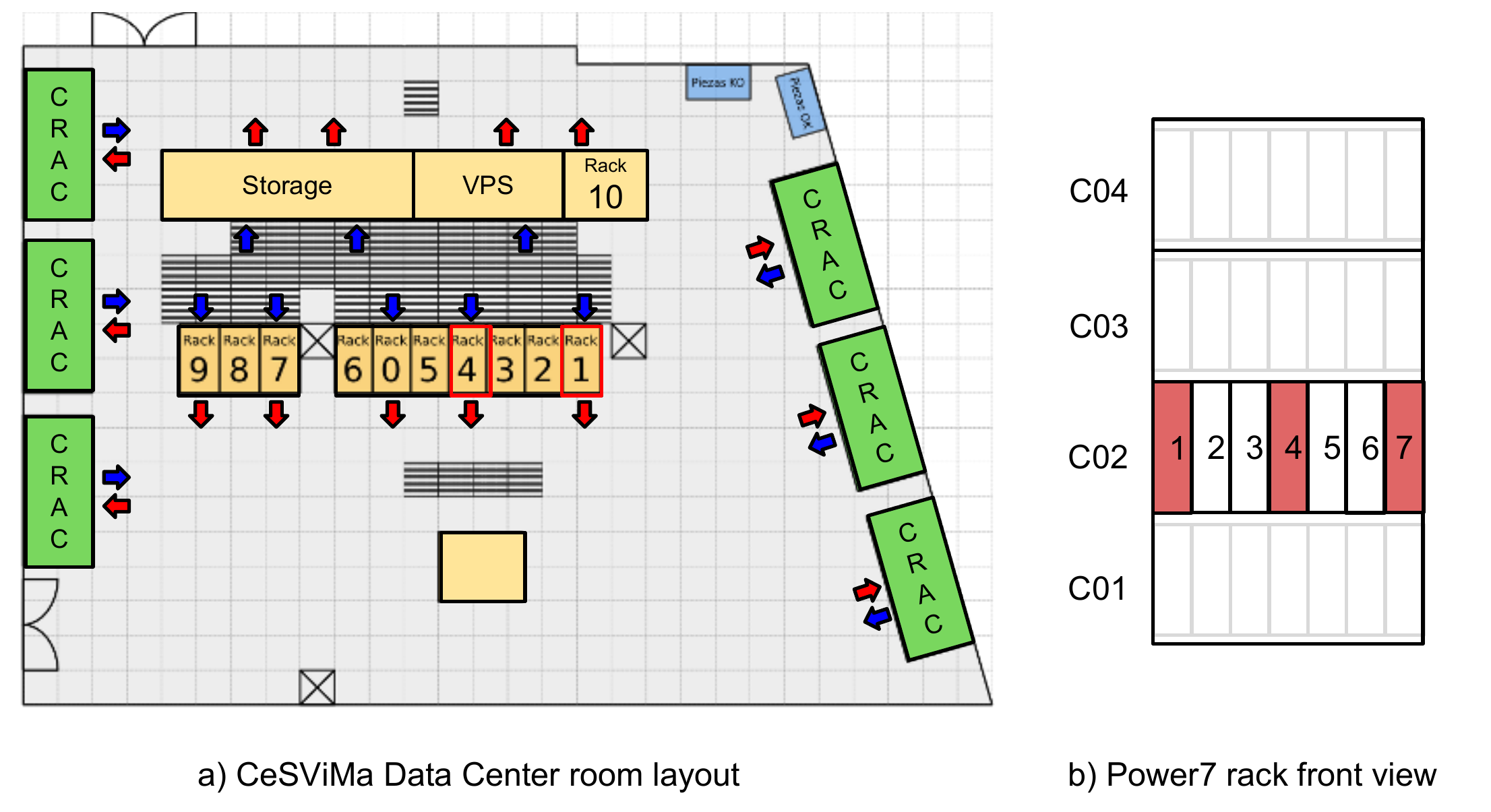}
  \caption{CeSViMa data room layout. Models are developed for
    Power7 nodes 1,4 and 7 at high c02 in racks 1 and 4.}
  \label{fig:cesvima-layout}
\end{figure}

CeSViMa data room has a cold-hot aisle layout and is cooled by means
of 6 CRAC units arranged in the walls that impulse air through the
floor plenum. To control data room cooling, the air return temperature
of each CRAC unit can be set independently. The room has a total size
of 190 square meters. Figure~\ref{fig:cesvima-layout}a shows the
layout of the data center. Rack 0 is a control rack that runs no HPC
computation. Racks 1-9 are filled with Power7 blade servers, whereas
rack 10 contains Intel Xeon servers. Each Power7 node is installed in
a blade center. Each blade center contains up to 7 blades, and each
rack contains 4 chassis (C01 to C04), as shown in
Figure~\ref{fig:cesvima-layout}b. To run our models, we have deployed
the same sensor network as in the reduced scenario. In particular,
to model inlet temperature we gather inlet temperature, humidity, CRAC
air return temperature and differential pressure through the floor
plenum. Because we have placed pressure sensors in the tiles in front
of racks 1 and 4, we model the Power7 nodes in these racks.  To model
CPU temperature, we also collect CPU temperature and fan speed of all
servers via IPMI. CeSViMa Power7 chassis do not have per-server power
sensors and we are not able to deploy current clamps. Thus, we use
per-server utilization as a proxy for the power consumption of the
node. 
As stated before, utilization is not an accurate metric for arbitrary
workloads. However, because of the nature of the workloads in CeSViMa
and only for thermal modeling purposes, utilization can be used as a
proxy variable to power consumption, 
as we show in Section~\ref{sec:results}.

In this work we show the modeling results for the servers highlighted
in red in Figure~\ref{fig:cesvima-layout}, i.e. nodes 1,4,7 at chassis
c02 of both rack 1 and 4. These nodes are the ones that exhibit the
most variable workload and extreme temperature conditions and
constitute the worst-case scenario for modeling.

\subsection{Modeling framework}
Because of the large number of servers in CeSViMa, to enable cooling
optimization we need a framework that allows to automatically model
and predict the CPU and inlet temperature of all servers. Even though
CeSViMa is a small-sized data room, it has a very high density in
terms of IT equipment. For instance, the amount of data gathered that
needs to be processed to enable full environmental modeling and
prediction, for a period of 1 year, is above 100GB. Thus,
modeling the whole data center with traditional approaches that
require human interaction is not feasible.

Our work uses the proposed GE techniques to
automatically model all the parameters involved in data center
optimization by automatically running the
training of the algorithms and testing them during runtime.  



%% file: results.tex
In this section we present the experimental results obtained when applying Grammatical Evolution to model CPU and inlet temperature. First, we show the results for the controlled scenario, describing the best algorithm configuration, and compare our method with state-of-the art solutions. Then, we apply the best configuration to train and test the models in a real data center scenario.

\subsection{Algorithm setup and performance}
First, we use GE to obtain a set of candidate solutions with low error
when compared to the temperature measurements in our controlled
experimental setup, under different constraints.

After evaluating the performance of our model with several setups, we
select the following one for each model in this paper:

\begin{itemize}
  \item Population size: 200 individuals
  \item Chromosome length: 100 codons
  \item Mutation probability: inversely proportional to the number of
    rules.
  \item Crossover probability: 0.9
  \item Maximum wraps: 3
  \item Codon size: 8 bits (values from 0 to 255)
  \item Tournament size: 2 (binary)
\end{itemize}

For CPU temperature prediction, we use a data window of $W_{cpu}=20$
samples (corresponding to 200 seconds) and a prediction window of
$\alpha=6$ (corresponding to 60 seconds). The data window is
heuristically chosen with respect to the largest observed temperature
transient, and its size is a trade-off between model accuracy and
convergence time. In this sense, the data window needs to incorporate
enough samples to capture the trends of thermal transients, but we
also need to consider that the larger the data window, the larger the
exploration space. The prediction window needs to be selected with
respect to the time it takes to actuate on the system and observe a
response. In our case, the data window size $W_{cpu}$ has been chosen
in accordance to our previous work on server power modeling, where we
analytically modeled temperature transients in enterprise servers,
observing that the largest transients, i.e. the worst-case modeling
scenario, occur for the lower fan speed
values~\cite{Zapater:TPDS:2014}. The prediction window is chosen given
the physical constraints of the problem: 1-minute prediction is
sufficient time to change the workload assignment of a server, as
canceling the workload of a server in case of thermal redlining takes
few seconds.

For inlet temperature prediction, we also use a data window of
$W_{inlet}=20$ samples but a prediction window of $\beta=5$
samples. Inlet temperature dynamics are much slower than CPU
temperature. Because of this, a sampling rate of 2 minutes over inlet
temperature is sufficient to get accurate results. Given the size of
the prediction window, we are able to obtain inlet temperature samples
10 minutes advance, which is sufficient time to act upon data
room cooling.

Next, we present the comparison among several configurations in terms
of grammar expressions and rules, premature convergence prevention and
fitness biasing. We detail our results for CPU temperature modeling.
The procedure to tune inlet temperature models is completely
equivalent.

\subsubsection{Data preprocessing and model simplification}
Because the power measurements of the Intel Xeon server are taken with
a current clamp, the power values obtained exhibit some noise. We
preprocess the data to eliminate high-frequency noise, smoothing
the power consumption trace by means of a low pass filter. The
remaining traces did not exhibit noise, so no preprocessing was
needed.

Moreover, we perform variable standardization for every feature (in the
range $[1, 2]$) to assure the same probability of appearance for all
the variables and to enhance the GE symbolic regression. 

\subsubsection{Grammars used}
To model CPU temperature we have tested three different grammars:
\begin{itemize}
\item The first is shown in Grammar~\ref{fig:bnf-cpu-basic} and
  contains a wide set of operands and preoperands (rules II and III),
  that do not necessarily yield models with a physical meaning.
\item The second grammar is a variation of
  Grammar~\ref{fig:bnf-cpu-basic} in which the number of preoperands
  (rule III) is reduced to exponentials only, i.e. $\synt{preop}::=
  exp$
\item The last grammar is the one presented in
  Grammar~\ref{fig:bnf-cpu} and also reduces the set of possible
  expressions (rule I).
\end{itemize}

\begin{algorithm}
\begin{numberedgrammar}
<expr> ::= <expr><op><expr>|(<expr><op><expr>)\\
| <preop>(<expr>)|<var>|<cte>

<op> ::= +|-|*|/

<preop> ::= exp | sin | cos | tan

<var> ::= TS[k-<idx>]|TIN[k-<idx>]|PS[k-<idx>]\\
|FS[k-<idx>]

<idx> ::= <dgt2><dgt>

<cte> ::= <dgt>.<dgt>

<dgt> ::= 0|1|2|3|4|5|6|7|8|9

<dgt2> ::= 0|1
\end{numberedgrammar}
\caption{Grammar used for CPU temperature modeling in BNF, that
  uses inlet temperature (TIN), fan speed (FS), power consumption
  (PS), past CPU temperature (TS) and past predicted CPU temperature
  (TpS)}
\label{fig:bnf-cpu-basic}
\end{algorithm}

From the previous three grammars the one that has faster convergence
time to achieve a low error, is Grammar~\ref{fig:bnf-cpu}.
Constraining the grammar improves convergence time and provides
phenotypes that have physical meaning, without an increase in the
modeling error obtained. Thus, for the remaining of the paper we work
with the simplified Grammar~\ref{fig:bnf-cpu} when modeling CPU
temperature.

\begin{algorithm}
\begin{numberedgrammar}
<expr> ::= <expr><op><expr>|(<expr><op><expr>)\\
 <preop>(<exponent>)|<var>|<cte>

<op> ::= +|-|*|/

<preop> ::= exp 

<exponent> ::= <sign><cte>*<var>\\
 |<sign><cte>*(<var><op><var>)

<sign> ::= +|-

<var> ::= TpS[k-<idx>]|TS[k-<idx>]|TIN[k-<idx>]\\
 |PS[k-<idx>]|FS[k-<idx>]

<idx> ::= <dgt>

<cte> ::= <dgt2>.<dgt2>

<dgt> ::= 1|2|3|4|5|6|7|8|9|10|11|12|13|14\\
 |15|16|17|18|19|20

<dgt2> ::= 0|1|2|3|4|5|6|7|8|9
\end{numberedgrammar}
\caption{Simplified grammar in BNF format used for CPU temperature
  modeling}
\label{fig:bnf-cpu}
\end{algorithm}

We test two variations of this grammar: i) one that searches for a
mixed model (i.e. uses past temperature predictions, and it is the one
shown in Grammar~\ref{fig:bnf-cpu}), and ii) the one that provides a
real model (i.e. only uses CPU temperature measurements). The only
difference between the mixed and the real grammars, is the presence of
the parameter $TpS$.

\subsubsection{Tested configurations}
With respect to premature convergence, we test three different
techniques:
\begin{itemize}
  \item No premature convergence technique applied
  \item Random Off-Spring Generation (2-RO) plus Packing, keeping no
    more that a 5\% of equal individuals.
  \item Random Off-Spring Generation (2-RO) plus Packing, leaving no
    more than 1 individual with equal phenotype. 
\end{itemize}

\begin{figure}
  \centering
  \includegraphics[width=0.9\columnwidth]{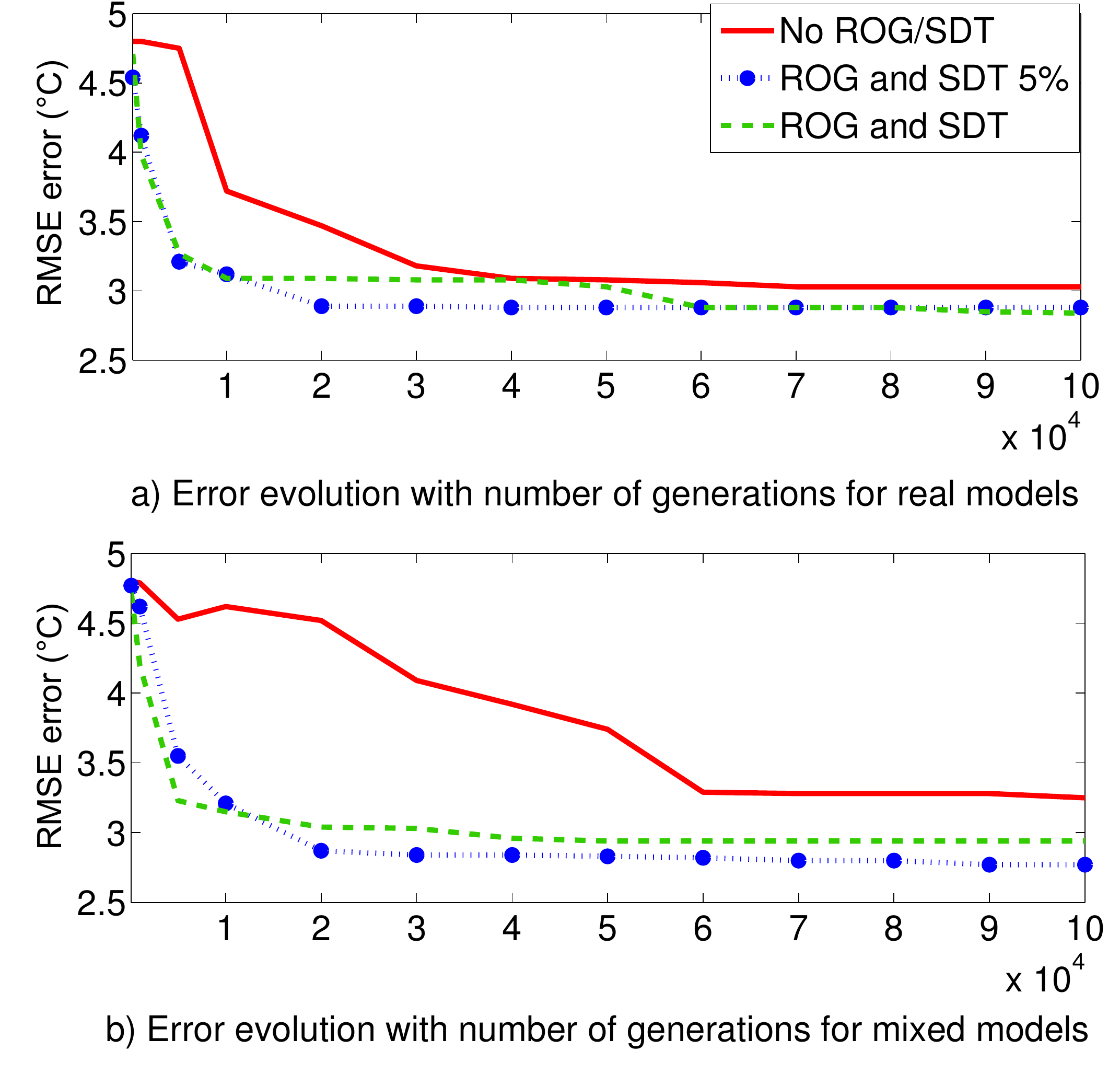}
  \caption{CPU temperature error evolution for real and mixed models
    under different premature convergence prevention techniques: i) no
    technique applied, ii) ROG + SDT keeping 5\% of equal individuals
    and iii) ROG + SDT randomizing all equal individuals.}
  \label{fig:error-evolution-sdt}
\end{figure}

For each of the previous configurations, we run both real and mixed
models, with the goal of comparing the convergence time and the fitness
evolution of each configuration. Because of the random evolution of
the algorithms, for comparison purposes, we run the same model
training 5 times and average the RMSE obtained for different number of
generations. Figure~\ref{fig:error-evolution-sdt} shows the RMSE
evolution for the three configurations, with both real and mixed
models. 

When we do not apply any technique, error decay is much
slower, as population loses diversity and improves only due to
mutation in the individuals. The impact is higher for the mixed
models, where search space is larger. When we apply ROG and SDT, we
need less generations to obtain good fitness values. However, keeping
only 1 individual with the same phenotype and randomizing the
remaining population is too aggressive, while keeping a higher
percentage of equal individuals, i.e. a 5\%, yields better results.
As shown, using 30,000 generations is enough to obtain low RMSE
values.

\begin{figure}
  \centering
  \includegraphics[width=0.9\columnwidth]{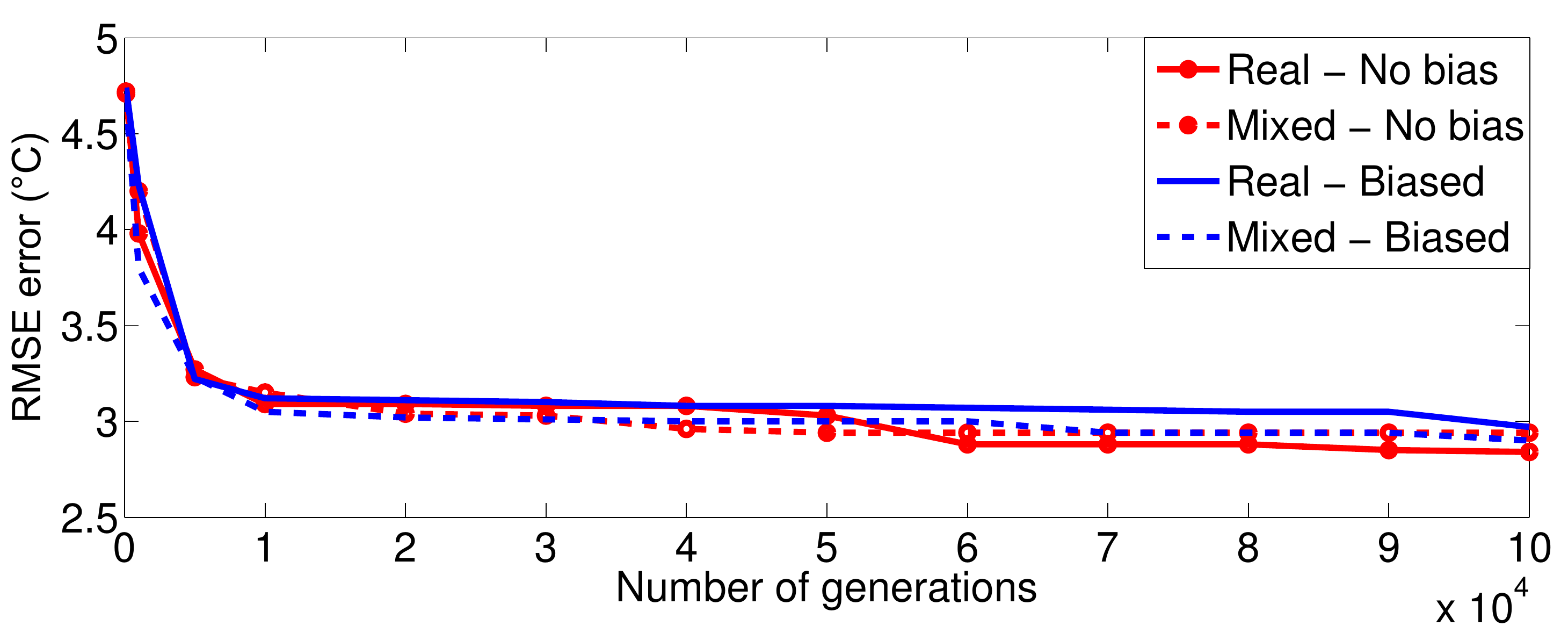}
  \caption{CPU temperature error evolution for real and mixed models
    under ROG + SDT 5\% when fitness is biased vs. not biased.}
  \label{fig:error-evolution-forced}
\end{figure}

Regarding fitness biasing, Figure~\ref{fig:error-evolution-forced} shows the
differences in terms of RMSE for different number of generations for
real and mixed models when we bias the fitness to force all parameters
and when we do not bias it. Convergence is similar, being slightly
better that of the non-biased models. In fact, all variables in the
grammar tend to appear in non-biased models, backing up the hypothesis
that all those magnitudes are correlated with temperature.

\begin{figure}
  \centering
  \begin{subfigure}[1-minute training set prediction for mixed model]
    {\includegraphics[width=1.0\columnwidth]{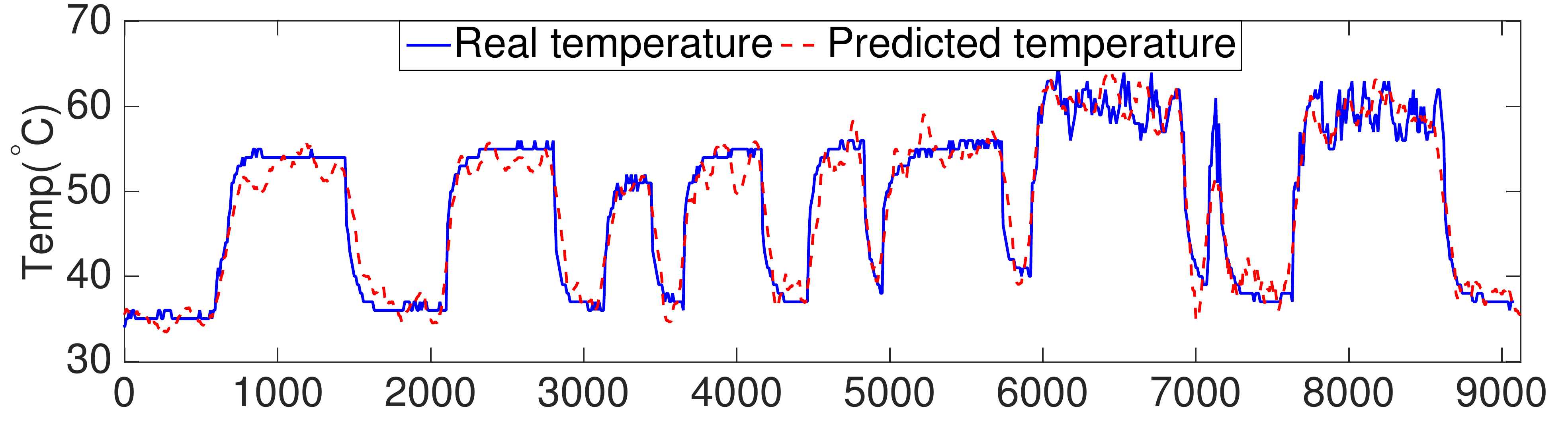}}
  \end{subfigure}%
  \begin{subfigure}[1-minute test set prediction for mixed model]
    {\includegraphics[width=1.0\columnwidth]{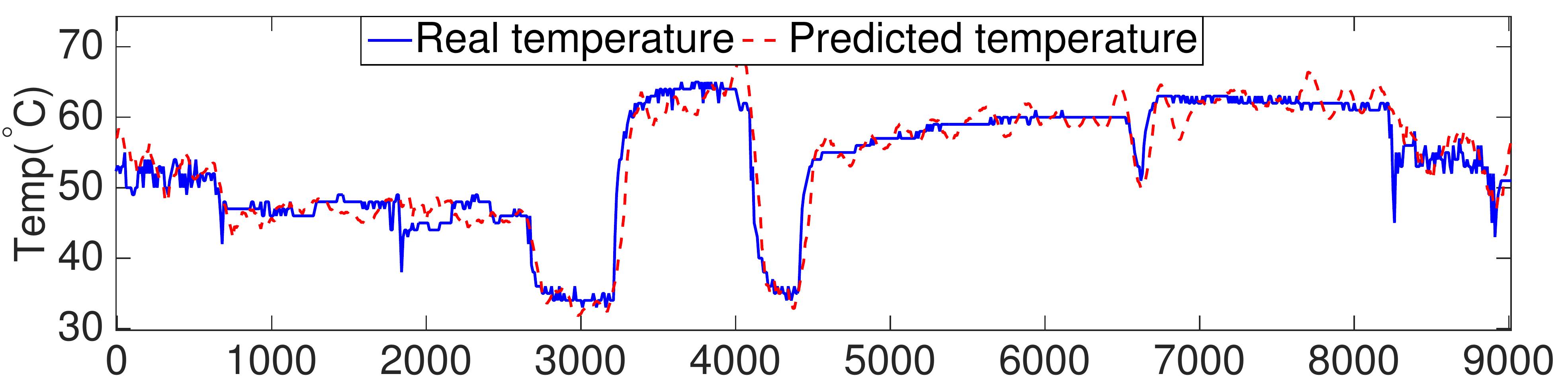}}
    \label{subfig:cpu-test}
  \end{subfigure}
  \caption{Training and test set CPU temperature prediction vs. real
    measurements with time (secs)}
  \label{fig:cpu-train-test}
\end{figure}

Finally, we show results for both the training and test set when
modeling CPU temperature with the best configuration, i.e. a mixed
model obtained with Grammar~\ref{fig:bnf-cpu}, using ROG and Packing
techniques leaving 5\% of equal individuals, and not biasing the
fitness. Table~\ref{tab:cpu-results} shows the 5 better phenotypes
obtained and their corresponding RMSE and MAE values for the test set
after simplification. To avoid overfitting, we use the five best
models to compute the samples of the test set, i.e., we predict the
next temperature sample with all five equations, obtaining 5 different
results, and we average them to obtain the prediction value. By
applying this methodology we obtain a RMSE of 2.48$\degree$C and a MAE
of 1.77$\degree$C. Because CPU temperature sensors usually have a
resolution of 1$\degree$C we consider these results to be accurate
enough for our purposes.  Figure~\ref{fig:cpu-train-test} shows a
zoom-in of the real CPU temperature trace and its prediction, for both
the training and the test set. As can be seen, the prediction
accurately matches the measured values in both the training and test
sets.

\begin{table*}[t]
\centering
\small
\begin{tabular}{lcc}
\hline 
Phenotype & RMSE & MAE\\ 
\hline\hline 
$TS[k-3]+PS[k-1]-PS[k-4]+1.1\cdot TIN[k-7]/(e^{(-0.1\cdot(TpS[k-8]-TS[k-3]))}$\\ 
$\cdot(PS[k-20]/9.9)+9.9)-(e^{(-4.5\cdot(TpS[k-5]/TS[k-19]))}\cdot TS[k-7])-9.6$
& 2.8 & 2.08\\ 
\hline
$TS[k-5]+PS[k-1])-PS[k-5]+(TS[k-5]/5.7-e^{(-1.3\cdot(TS[k-5]/TIN[k-12]))}\cdot
TS[k-5])$\\
$+PS[k-11]/(3.7-TS[k-5]\cdot(e^{(-0.3\cdot(PS[k-18]+FS[k-7]))}-e^{(-4.9\cdot(PS[k-12]/PS[k-16]))}))-9.8$
& 2.59 & 1.86\\
\hline
$TS[k-5]+PS[k-1]-PS[k-4]+e^{(+6.1\cdot(TIN[k-10]/TS[k-5]))}-5.2)-TIN[k-14]\cdot(0.06\cdot
(TIN[k-7]/TS[k-5]))$\\
$+(PS[k-1]/3.1\cdot TIN[k-7])\cdot TS[k-5]-(PS[k-20]/TIN[k-7])\cdot^{(-6.4\cdot(TIN[k-15]/TS[k-11]))}-8.0$
& 2.77 & 2.01\\
\hline
$TS[k-3]+PS[k-1]-PS[k-4]-e^{(-4.1\cdot(TpS[k-5]/TIN[k-19]))}/TS[k-3]$\\
$+(TpS[k-1]/(e^{(-0.1\cdot(TpS[k-8]-TS[k-3]))}\cdot(PS[k-20]/9.9)+9.2))/1.6-9.6$
& 2.55 & 1.77\\
\hline
$TS[k-1]+0.73\cdot(PS[k-1]\cdot 4.4-PS[k-2]\cdot
4.4-TS[k-1]+TpS[k-8]+e^{(-0.2\cdot TIN[k-8])}$\\
$+(e^{(-3.4\cdot(PS[k-20]/PS[k-10]))}-e^{(-4.0\cdot(PS[k-6]/PS[k-19]))})\cdot TpS[k-12]/9.0-0.9)$
& 2.5 & 1.75 \\
\hline
\end{tabular}
\caption{Phenotype, RMSE and MAE for the test set in the CPU
  temperature modeling reduced scenario}
\label{tab:cpu-results}
\end{table*}

\subsection{Comparison to other approaches}
We compare our results with three common techniques for CPU temperature modeling in the state-of-the-art: autoregressive moving average models, linear subspace identification techniques and dynamic neural networks. We first briefly describe these three modeling techniques and then we show the results obtained and compare them with our proposed technique.

\subsubsection{ARMA models}
ARMA models are mathematical models of autocorrelation in a time
series, that use past values alone to forecast future values of a
magnitude. ARMA models assume the underlying model as stationary and
that there is a serial correlation with the data, something that
temperature modeling accomplishes. In a general way, an ARMA model can
be described as in Equation~\ref{eq:arma}:

\begin{equation}
y_t = \sum_{i=1}^{p}(a_i \cdot y_{t-i}) = e_t + \sum_{j=1}^{q}(c_j \cdot e_{t-j})
\label{eq:arma}
\end{equation}

\noindent where $y_t$ is the value of the time series (CPU temperature
in our case) at time $t$, $a_i$'s are the lag-i autoregressive
coefficients, $c_i$'s are the moving average coefficient and $e_t$ is
the error. The error is assumed to be random and normally
distributed. $p$ and $q$ are the orders of the autoregressive (AR) and
the moving average (MA) parts of the model, respectively.

The ARMA modeling methodology consists on two different steps: i)
identification and ii) estimation. In our work we use an automated
methodology similar to the one proposed by Coskun
et.al.~\cite{Coskun:TCAD:2009}. During the identification phase, the
model order is computed, i.e, we find the optimum values for $p$ and
$q$ of the $ARMA(p,q)$ process. To perform model identification we use
an automated strategy, that computes the goodness of fit for a range
of $p$ and $q$ values, starting by the simplest model (i.e., an
ARMA(1,0)). The goodness of fit is computed using the Final Prediction
Error (FPE), and the best model is the one with lowest FPE value,
given by Equation~\ref{eq:fpe}:

\begin{equation}
FPE=\frac{1+n/N}{1-n/N}\cdot V
\label{eq:fpe}
\end{equation}

\noindent where $n=p+q$, $N$ is the length of the time series and $V$
is the variance of the model residuals. For a fair comparison with our
proposed methodology, the model obtained needs to forecast $\alpha=6$
samples.

\subsubsection{N4SID}
N4SID is a subspace identification method that estimates an $n$ order
state-space model using measured input-output data, to obtain a model
that represents the following system:
\begin{eqnarray}
\dot{x}(t) = Ax(t) + Bu(t)+Ke(t)\\
y(t) = Cx(t) + Du(t) + e(t)
\end{eqnarray}

\noindent where $A$,$B$,$C$ and $D$ are state-space matrices, $K$ is
the disturbance matrix, $u(t)$ is the input, $y(t)$ is the output,
$x(t)$ is the vector of $n$ states and $e(t)$ is the disturbance.

State-space models are models that use state variable observations to
describe a system by a set of first-order differential equations,
using a black-box approach. The approach consists on identifying a
parameterization of the model, and then determining the parameters so
that the measurements explain the model in the most accurate possible
way.  They have been very successful for the identification of linear
multivariable dynamic systems.

To be constructed, certain parameters need to be fed into the model,
such as the number of forward predictions ($r$), the number of past
inputs ($s_u$), and the number of past outputs($s_y$).  Again, for a
fair comparison with our proposed methodology, we need a model in the
form $N4SID[r,s_u,s_y]$ where $r=6, s_u=20$ and $s_y=20$.

\subsubsection{NARX}
A Nonlinear Autoregressive eXogenous (NARX) model is a nonlinear autoregressive model with exogenous inputs. In this case, the current value of a time series is computed in function of a) past values of the same series, and b) current and past values of the exogenous series. Additionally, the model contains an error term, since knowledge of the other terms will not enable the current value of the time series to be predicted with precision. This is model is described as follows:

\begin{equation}
y(t) = F(y(t-1). y(t-2), y(t-3), \ldots, u(t), u(t-1), u(t-2), u(t-3), \ldots) + e(t)
\end{equation}

\noindent where $y(t)$ is the output, $u(t)$ is the exogenous variable and $e(t)$ is the error term. $F$ is a nonlinear function, and in our case is defined through a neural network. The NARX model is based on the linear ARX model, which is commonly used in time-series modeling.

\subsubsection{Model comparison}
Finally, we compare the results for CPU temperature modeling between our proposed approach, ARMA, N4SID, and NARX models, all of them with a prediction window of 6 samples (1 minute). To perform a statistical comparison, we have conducted a non-exhaustive 5-fold cross-validation. The complete data set is obtained from more than 10 hours of temperature traces gathered from an Intel Xeon RX-300 S6 server running a wide range of workloads under various cooling setups. This data set has been partitioned into 5 equal sized subsamples. A single subsample is retained as the validation data for testing the model, and the remaining 4 subsets are used as training data. This process is repeated 5 times with each of the 5 subsamples used exactly once as the validation data. RMSE and MAE are averaged to perform a comparison between ARMA, N4SID, NARX and GE. Each resultant set of five models will be averaged to produce the final estimation.

\begin{table}[t]
\centering
\footnotesize
\begin{tabular}{lcccc}
\hline
Model & \multicolumn{2}{c}{Training set} & \multicolumn{2}{c}{Test set} \\
      & RMSE & MAE & RMSE & MAE\\ 
\hline
ARMA$_{1,4}$   & $3.38 \pm 0.23$ & $1.62 \pm 0.13$ & $3.23 \pm 1.02$ & $1.62 \pm 0.57$ \\ 
ARMA$_{9,8}$   & $3.35 \pm 0.24$ & $1.70 \pm 0.14$ & $3.24 \pm 1.01$ & $1.74 \pm 0.64$ \\ 
N4SID       & $2.55 \pm 0.28$ & $1.69 \pm 0.07$ & $3.98 \pm 1.11$ & $2.93 \pm 0.58$ \\ 
NARX        & $2.53 \pm 0.10$ & $1.64 \pm 0.05$ & $3.88 \pm 1.16$ & $2.35 \pm 0.69$ \\ 
GE          & $2.32 \pm 0.19$ & $1.50 \pm 0.08$ & $2.56 \pm 0.90$ & $1.66 \pm 0.45$ \\
 
\hline
\end{tabular}
\caption{RMSE and MAE in CPU temperature prediction for each model 
  (ARMA, N4SID, NARX and GE)}
\label{tab:model-comp}
\end{table}

\begin{figure}
  \centering
  \includegraphics[width=0.95\columnwidth]{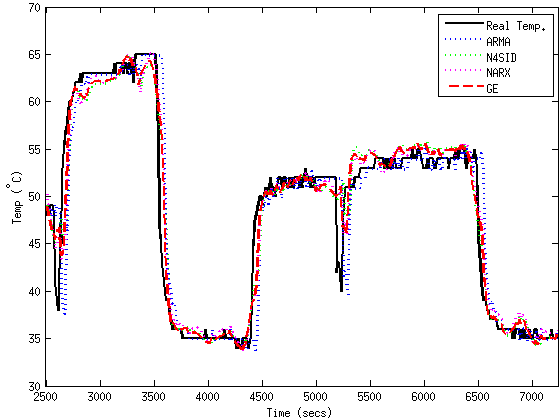}
  \caption{Zoomed-in averaged CPU temperature modeling comparison between
    ARMA(1,4), N4SID, NARX and GE}
  \label{fig:model-comp}
\end{figure}

Table~\ref{tab:model-comp} shows the RMSE and MAE errors obtained for our proposed modeling technique based on GE, ARMA, N4SID and NARX models, and Figure~\ref{fig:model-comp} shows a zoom-in into the CPU temperature curve for the actual measurements and the averaged prediction of the five models obtained in the cross validation over the whole data set. As can be seen, GE models are the ones with both lower RMSE and MAE. Moreover, the CPU temperature trend is accurately predicted. This does not happen for ARMA models that, even though keep the maximum error low, provide values that are always behind the real trend, yielding poor forecasting capabilities. This issue cannot be solved by increasing the model order, as shown in Table~\ref{tab:model-comp}. N4SID models, even though they are very accurate in the training set, perform poorly in the test set and have an important bias error. Even if the bias error is corrected (which has been done in Figure~\ref{fig:model-comp}) the prediction is still behind the measurements and the model is unable to capture the system dynamics. NARX models clearly show an overfitting effect in the training data. The GE prediction, even though has more oscillations (due to the smoothed noise of the power consumption signal) is the only one that captures the temperature trend, advancing the real measurements.

\subsection{Inlet temperature modeling}

For inlet temperature modeling we perform the same study than for CPU
temperature in terms of grammars, premature convergence and
fitness biasing. As expected, the results in terms of the best model
configuration yield very similar results. Thus, for inlet temperature
modeling, we use the same configuration: i) a mixed model using a
simplified version of the grammar that only allows exponentials, ii) SDT
with 5\% packing and iii) RMSE fitness function without biasing.

\begin{figure}
  \centering
  \includegraphics[width=1.0\columnwidth]{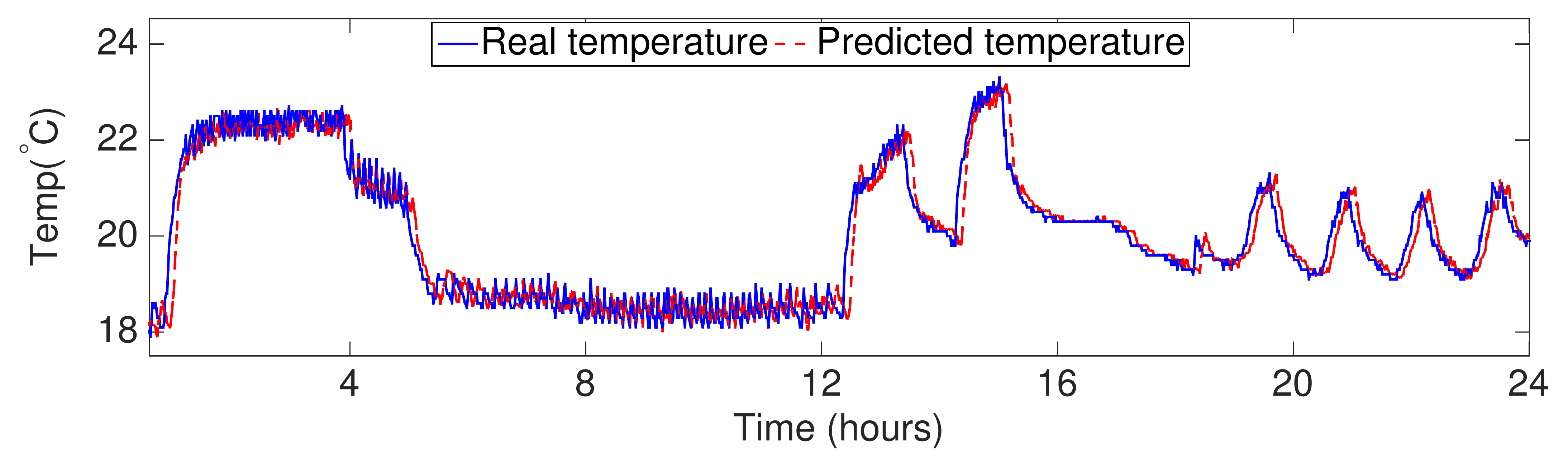}
  \caption{10-minute inlet temperature prediction in the reduced
    scenario for a mixed model with SDT Packing 5\% and simplified grammar}
  \label{fig:tinlet-test}
\end{figure}

The BNF grammar used is very similar to Grammar~\ref{fig:bnf-cpu},
where instead of rule VI, we use the following rule:

\begin{grammar} 
<var> ::= TIN[k-<idx>] | TpIN[k-<idx>] 
$\alt$ TSUP[k-<idx>] | HUM[k-<idx>] 
\end{grammar}

\noindent where TSUP is cold air supply temperature, HUM is humidity,
TIN are past inlet temperatures and TpIN are past inlet temperature
predictions. Figure~\ref{fig:tinlet-test} shows the prediction for the
test set. The RMSE of the prediction is of 0.33$\degree$C and MAE is
0.27$\degree$C for a prediction window of 10 minutes
and for the test set. Again, the model includes all the available
variables, i.e., TSUP, TIN and HUM appear in the final model.

\subsection{Data Center modeling}
We use the previous model with the same configuration to predict the
CPU and inlet temperature of the blade servers at CeSViMa data
center. Because CeSViMa is a production environment, when it comes to
server data, we are subject to the data sampling rates provided by the
data center. CeSViMa collects all data from servers every 2 minutes,
and environmental data (i.e. from coolers) every 15 minutes. Thus, for
both CPU and inlet temperatures, we need to change our prediction
windows.  For CPU temperature we use a prediction window $\alpha '=1$,
which means that we predict CPU temperature two minutes into the
future.  For inlet temperature we use a prediction window $\beta '=1$
samples, i.e. we predict temperature 15 minutes ahead.

Because in CeSViMa we cannot control the workload being executed, nor
modify the cooling setup, we need to select longer training and tests
sets to ensure that they exhibit high variability on the magnitudes of
interest. For CPU temperature, we select 2 days of execution for the
training set, and 4 days for the test set. For inlet temperature,
which varies very slowly in a real data center setup we use 14 days of
execution for both the training and the test set.

\begin{figure}
  \centering
  \includegraphics[width=1.0\columnwidth]{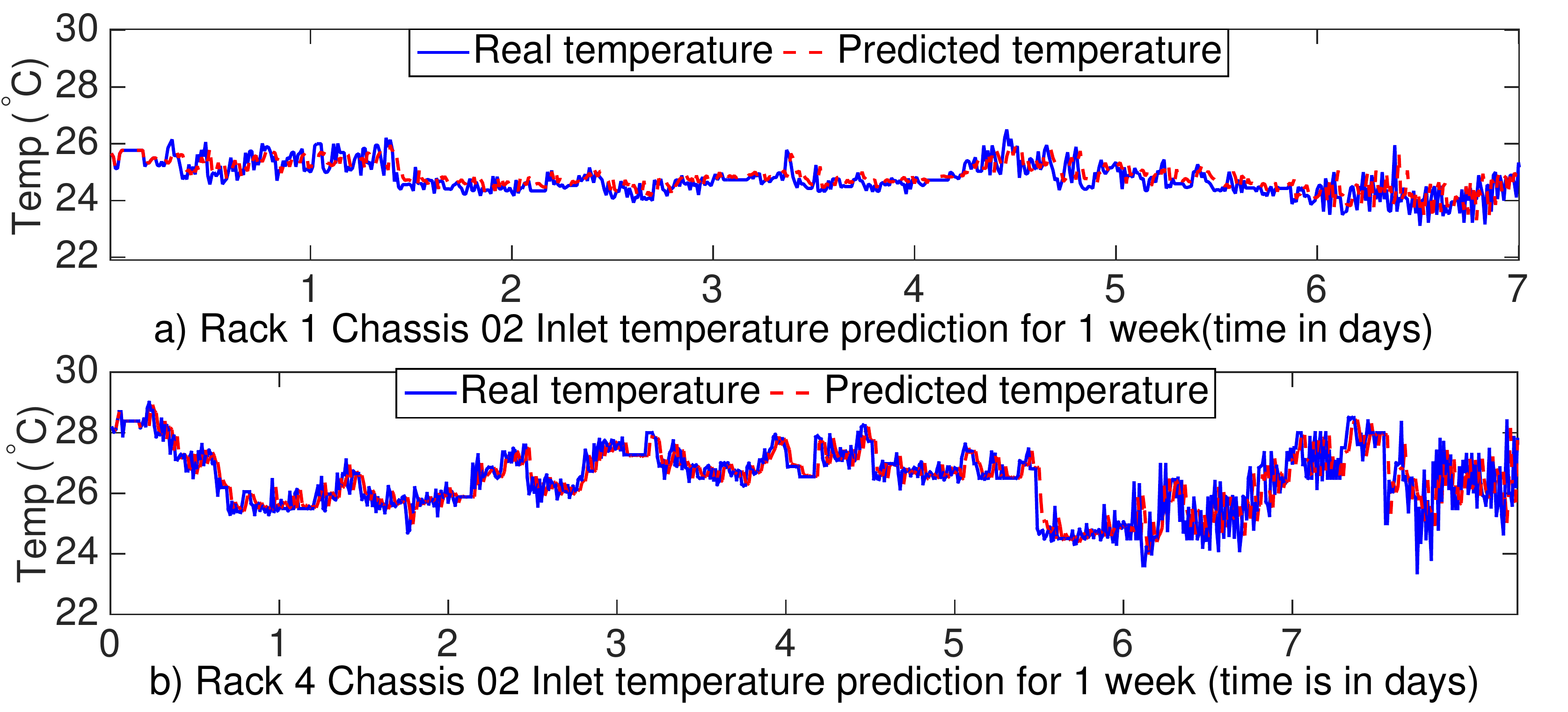}
  \caption{Inlet temperature modeling for various racks}
  \label{fig:cesvima-tinlet}
\end{figure}

\begin{figure*}
  \centering
  \includegraphics[width=1.0\textwidth]{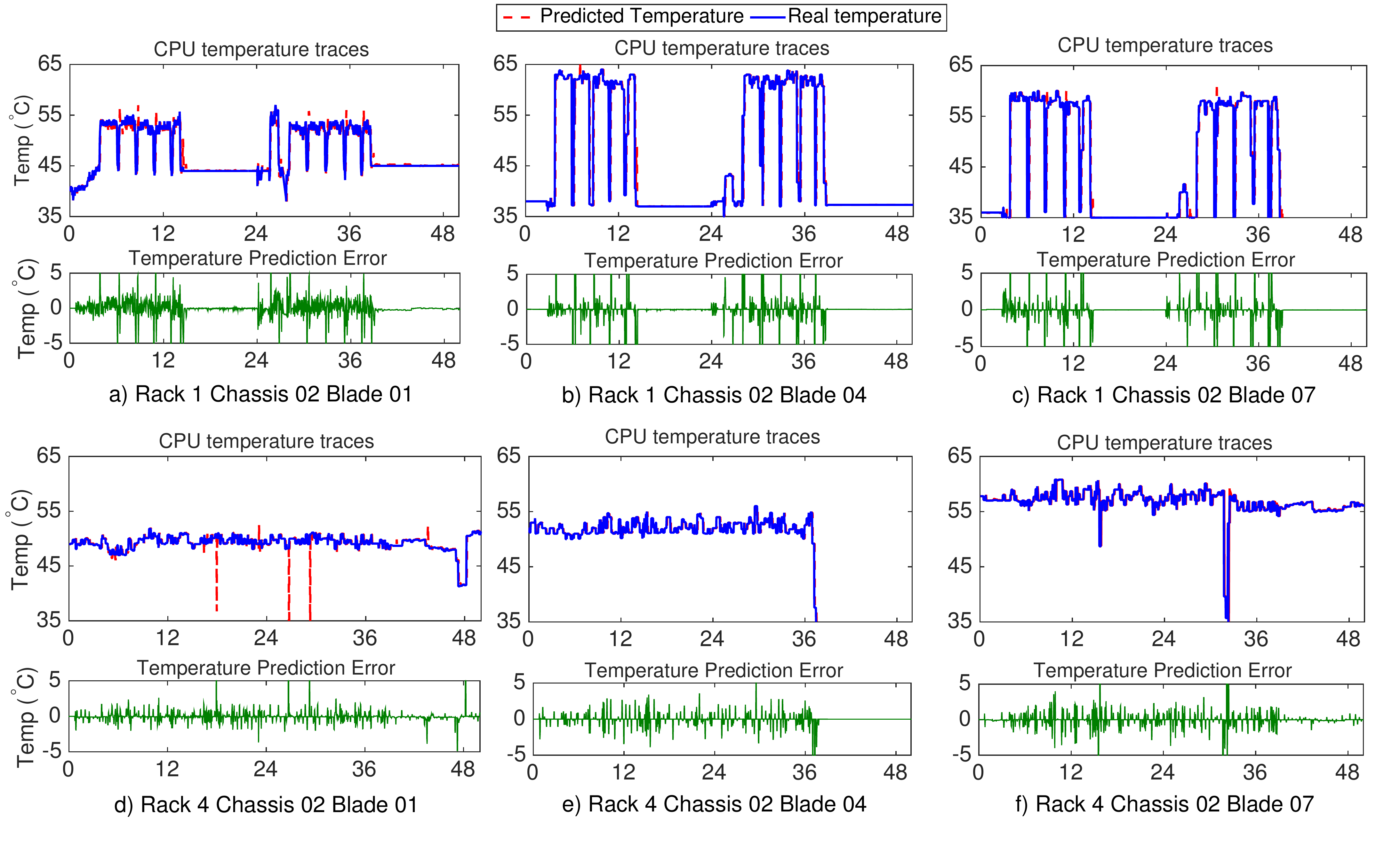}
  \caption{Data Center CPU temperature modeling and prediction error
    for various servers in different racks for 2 days of traces (time
    in hours)}
  \label{fig:cesvima-tcpu}
\end{figure*}

Figure~\ref{fig:cesvima-tinlet} shows a zoomed-in plot of the measured
and predicted inlet temperature to the chassis c02 of racks 1 and 4 in
CeSViMa data center for a period of 8
days. Figure~\ref{fig:cesvima-tcpu} shows the measured and predicted
CPU temperature traces for blades b01, b04 and b07 in chassis c02 of
both racks, for the first two days of the same period, as well as the
prediction error (i.e. the difference between the real measurements
and the prediction). To generate these last models, instead of using
the real inlet temperature measurements, we use predicted inlet
temperature. This way, we are able to accurately predict all variables
needed for optimization.

\begin{table*}[t]
\centering
\begin{tabular}{llcc}
\hline 
Model & Phenotype & Train. & Test\\ 
\hline\hline 
Inlet &  
$TIN[k-1]+e^{(-4.3\cdot (TRET2[k-3]/TRET2[k-11]))}/TIN[k-1]\cdot
(e^{(+1.5\cdot TIN[k-5])}$ & 0.32 & 0.4 \\
Rack1 & $-e^{(-3.8*(PDIF[k-20]-TIN[k-1]))})$ 
&  & \\ 
\hline
Inlet &
$TIN[k-1]+3.1/e^{(+5.0\cdot TIN[k-1])}\cdot
e^{(+2.2\cdot TIN[k-2])}-e^{(-2.9\cdot TpIN[k-3])}\cdot TRET3[k-12]\cdot $ & & \\
Rack4 &
$(TRET3[k-5]+e^{(-4.9\cdot
  (HUM[k-6]/TRET2[k-1]))}/TIN[k-2]/e^{(+7.2\cdot (PDIF[k-20]-TIN[k-1]))})$
& 0.18 & 0.44 \\
\hline\hline
CPU &
$TS[k-1]+(PS[k-20] \cdot FS[k-1]-9.4 \cdot (TpS[k-5] \cdot
TpS[k-2]))/(e^{(+2.0 \cdot FS[k-7])}/$ & & \\
Rack1, C02, B01 &  
$e^{(-4.1 \cdot TpS[k-5])}/(2.3+
e^{(+1.7*(TpS[k-10]*TpS[k-10]))}+ e^{(+1.5*FS[k-1])}-
e^{(+1.6*PS[k-7])}))$ & 0.68 & 0.76 \\
\hline
CPU & 
$TS[k-1]+e^{(-7.2*(TS[k-6]/PS[k-1]))}/(e^{(-6.1*(TS[k-10]-PS[k-15]))}/5.6/FS[k-20]$ & &\\
Rack1, C02, B04 &
$-1.7+TpS[k-20]-e^{(-3.0*(TIN[k-4]/TS[k-15]))})$
&  & \\
\hline
CPU &
$TS[k-1]+e^{(-6.2\cdot (TS[k-2] \cdot TIN[k-11]))}/((e^{(-9.8 \cdot
  TS[k-8])} + e^{(-5.2*(TS[k-9]*PS[k-19]))}) $ & 0.51 & 0.85 \\
Rack1, C02, B07 &
$\cdot e^{(+5.8*FS[k-9])})$
&  &  \\
\hline
CPU &
$TS[k-1]+e^{(-3.1\cdot TS[k-2])}/(((TS[k-3]/PS[k-3])
+(FS[k-18]-FS[k-8])/$ & 0.55 & 0.75 \\
Rack4, C02, B01 &
$e^{(-9.4\cdot (FS[k-1]-TpS[k-9]))})-TpS[k-4]+(TpS[k-6]+TS[k-3])-
(TIN[k-3]/TpS[k-9]))$
&  0.29 & 0.46 \\
\hline
CPU \\ Rack4, C02, B04 &
$TS[k-1]+e^{(-5.7\cdot (TpS[k-6]*TpS[k-10]))}/e^{(+9.9\cdot (TpS[k-9]-TIN[k-10]))}$
&  0.26 & 0.73 \\
\hline
CPU \\ Rack4, C02, B07 &
$TS[k-1]+TIN[k-11]\cdot e^{(-9.5\cdot
  (TpS[k-10]/TIN[k-5]))}/e^{(-9.9\cdot (TIN[k-10]-TS[k-5]))}$ 
& 0.43 & 0.87 \\

\hline
\end{tabular}
\caption{Phenotype and average error (in Celsius) in training and test
  set for CPU and inlet temperature modeling in a production Data
  Center}
\label{tab:cesvima}
\vspace{-0.2cm}
\end{table*}

Table~\ref{tab:cesvima} shows the phenotypes obtained for CPU and
temperature modeling of the servers in CeSViMa data
center. We also report MAE for both training and test
sets. We observe that all phenotypes that model CPU temperature
incorporate the parameters of interest (inlet temperature, power and
fan speed), and we obtain errors below 1$\degree$C in all cases. The
average RMSE across models are 1.52 $\degree$C and 1.57$\degree$C for
the training and test set respectively.  As for inlet temperature, the
phenotypes incorporate both differential pressure, and CRAC return
temperature. Moreover, depending on the rack placement, the influence
of the CRAC units vary. Here we can observe the benefits of the
feature selection performed by GE. Rack1, which is the leftmost rack
in the data center, is affected only by CRAC2; whereas Rack4, situated
in the middle of the row, is affected both by CRAC2 and CRAC3. The
model automatically incorporates the most relevant features,
discarding the irrelevant ones. For inlet temperature prediction, our
error is below 0.5$\degree$C, which is enough for our purposes and
below other state-of-the-art approaches.


%% file: discussion.tex

%


In this section we briefly discuss the applicability of our models,
and the computational effort needed to model a full
data center scenario, to validate the feasibility of our approach.

\subsection{Applicability}

The goal of our modeling is predicting server CPU temperature under
variable cooling setups, so that cooling-associated costs can be
reduced without incurring on reliability issues. To this end, we first
predict the inlet temperature of servers given the data room
conditions and cooling setup, and use this result to predict server
temperature.

Having analyzed the spatio-temporal variability of inlet temperature
traces in CeSViMa data center, we find that it is sufficient to
predict inlet temperature at 3 different heights (at the bottom,
middle and top of the rack), in one out of two racks. This way, we
need to generate 30 inlet temperature models at most. Because the
maximum CPU temperature in the data center is the one limiting the
cooling, at most we need to predict the CPU temperature of each server
in the data room, i.e. we need as many models as servers in the data
center. However, if by analyzing the traces we find that there is a
subset of CPUs that limit the maximum cooling of the overall data
center, our problem can be reduced to modeling those that always
exhibit higher temperatures. For the particular case of the traces of
CeSViMa, if we examine 6 months of CPU temperature traces, we find
that the CPUs limiting the cooling are the blades b04 and b07 placed
in the second chassis (c02) of all racks. In this sense, for energy
optimization purposes our problem reduces to generating 10 different
models.
These models allow us to predict the maximum server temperature
attained in the data center and, thus, detect any potential thermal
redlining and act before it occurs.
Moreover, to leverage energy optimization, our results can be used to
set cooling dynamically during runtime, by predicting the maximum data
center CPU temperature under various cooling conditions and increasing
CRAC air supply temperature without incurring in reliability issues.

Even though in this paper we have applied our modeling methodology to
a raised-floor air-cooled data center scenario, the proposed technique
is also valid for data centers equipped with other state-of-the-art
cooling mechanisms, such as in-row or in-rack cooling used in
high-density racks.

\subsection{Computational effort}
Our approach is computationally intensive in the model training
stage. The GE model needs to evolve a random initial population for
30,000 generations to obtain accurate results. In our experiments,
running 30,000 generations of 4 different models in parallel takes 28h
in a computer equipped with a QuadCore Intel i7 CPU @3.4GHz and 8GB of
RAM. This computational cost is much larger than the computational
cost for training ARMA and N4SID models. However, to obtain accurate
results ARMA needs to be manually tuned, and N4SID requires a manual
feature selection step that greatly impacts accuracy, whereas GE
models can be automatically developed.

However, as the models obtained for homogeneous servers
are very similar, it is possible to reduce the training overhead by
using already evolved populations to fine-tune the models instead of
using the a new random population every time. This way, we can reduce
the training time significantly. 

As for the model testing, in the worst case scenario, the model needs
to be tested every 10 seconds. The overhead to test one model is found
to be negligible. In this sense, it is feasible to compute all
temperatures to find the maximum. Moreover, because of the temperature
imbalances of servers in the data room we can reduce the amount of
models run to those that are limiting the cooling, i.e. the servers
with higher CPU temperature values. Overhead incurred by testing is in
the same order of magnitude than the overhead of ARMA and N4SID, but
provides better results in terms of error.



%% file: conclusions.tex
In this paper we have presented a methodology for the unsupervised
generation of models to predict on runtime the thermal behavior of
production data centers running arbitrary workloads and equipped with
heterogeneous servers. 

Our approach leverages the usage of Grammatical Evolution to
automatically generate models of the data room by using real data
center traces. Our solution allows to predict the CPU temperature and
inlet temperature of servers, with an average error below 2$\degree$C
and 0.5$\degree$C respectively. These errors are within the margin
obtained by other off-line supervised approaches in the state-of-the
art. Our solution, generates the models in an unsupervised way,
is able to work on runtime, is trained and tested in
a real scenario, and does not require the usage of CFD software.

To the best of our knowledge our work is the first to propose data center
temperature forecasting using evolutionary techniques, allowing predictive model generation for runtime optimization.



%% file: appendix.tex
In this Appendix we provide further information on the mapping process
used by our grammar. For a more detailed explanation on the principles
of GE, the reader is referred to~\cite{Ryan:LBPGP:1998}. A BNF
specification is a set of derivation rules, expressed in the form:

\begin{grammar}
<symbol> ::= <expression>
\end{grammar}

Rules are composed of sequences of terminals, which are items that
can appear in the language, and non-terminals, which can be expanded
into one or more terminals and non-terminals. 
A grammar is represented by the tuple ${N, T, P, S}$, being $N$ the
non-terminal set, $T$ is the terminal set, $P$ the production rules
for the assignment of elements on $N$ and $T$, and $S$ is a start
symbol that should appear in $N$. The options within a production rule
are separated by a ``$|$'' symbol.

\begin{algorithm}
N = \{expr, op, preop, var, num, dig\}\\
T = \{+, -, *, /, sin, cos, exp, x, y, z, 
     0, 1, 2, 3, 4, 5, (, ), .\}\\
S = \{expr\}\\
P = \{I, II, III, IV, V, VI\}\\
\begin{numberedgrammar}
<expr> ::= <expr><op><expr> | <preop>(<expr>) | <var>

<op> ::= +|-|*|/

<preop> ::= sin|cos|log

<var> ::= x|y|z

<num> ::= <dig>.<dig> | <dig>

<dig> ::= 0 | 1 | 2 | 3 | 4 | 5
\end{numberedgrammar}
\caption{Example of a grammar in BNF designed for symbolic
  regression}
\label{fig:BnfExample}
\end{algorithm}

Grammar~\ref{fig:BnfExample} represents an example grammar in BNF. The
final expression consists of elements of the set of terminals $T$,
which have been combined with the rules of the grammar.

The chromosome is used to map the start symbol onto terminals by
reading genes (or codons) of 8 bits to generate a corresponding
integer value, from which the options of a production rule are
selected by using the modulus operator:

\begin{equation}
\text{Rule} = \text{Codon Value \% Number of Rule Choices} 
\label{eq:mapping}
\end{equation}

\noindent\textbf{Example:} In this example, we explain the mapping
process performed in GE to obtain a phenotype (mathematical function)
given a genotype (chromosome). Let us suppose we have the BNF grammar
provided in Figure~\ref{fig:BnfExample} and the following 7-gene
chromosome:

\begin{small}
\begin{verbatim}
21-64-17-62-38-254-2
\end{verbatim}
\end{small}

According to Figure~\ref{fig:BnfExample}, the start symbol is
$S=\synt{expr}$, hence the decoded expression begins with the
non-terminal:

\begin{equation*}
Solution=\synt{expr}
\end{equation*}

Now, we use the first gene of the chromosome (i.e. 21) in rule
\texttt{I} of the grammar. The number of choices in that rule is
3. Hence, the mapping function is applied: \verb|21 MOD 3 = 0| and the
first option is selected \synt{expr}\synt{op}\synt{expr}. The selected
option substitutes the decoded non-terminal, giving the following
expression:

\begin{equation*}
Solution=\synt{expr}\synt{op}\synt{expr}
\end{equation*}

The process continues with the codon 64, used to decode the first
non-terminal of the current expression, \synt{expr}. Again, the
mapping function is applied to the rule: \verb|64 MOD 3 = 1| and the
second option $\synt{preop}(\synt{expr})$ is selected. So far, the
current expression is:

\begin{equation*}
Solution = \synt{prep}(\synt{expr})\synt{op}\synt{expr}
\end{equation*}

The next codons (17, 62, 38, 254 and 2) are decoded in the same way. After
codon 2 has been decoded, the expression is:

\begin{equation*}
Solution = exp(x)*\synt{var}
\end{equation*}

At this point, the decoding process has run out of codons, and we need
to reuse codons starting from the first one. This technique is known
as wrapping and mimics the gene-overlapping phenomenon in many
organisms~\cite{Hemberg:GPEM:2013}. Applying wrapping, we use gene 21
to decode \synt{VAR} with rule \texttt{IV}. This result gives the
final expression of the phenotype:

\begin{equation*}
Solution = exp(x)*y
\end{equation*}

Apart from performing parameter identification, in conjunction with a
well-defined fitness function, the evolutionary algorithm is also
computing mathematical expressions with the set of features that best
fit the target system. Thus, GE is also defining the optimal set of
features that derive into the most accurate model.

\noindent\textbf{Adding time dependence:} Previously shown grammars
allow us to obtain phenotypes that depend on a certain number of
variables (e.g. $x,y,z$). We could use the previous method to predict
variables that depend only on the current observation of other
magnitudes, such as server power~\cite{Arroba:JGC:2014}.

Models created this way can be used to predict magnitudes without
memory and the data used for model creation consists of
samples. Temperature, however, is a magnitude with memory, i.e. the
current temperature depends on past temperature values. Thus, the data
used for model creation need to be a time series. By properly tuning
our grammars, we can add time dependence to the variables in the
phenotype, so that past values can be used to predict the variable a
certain number of samples ahead.

%% file: main.bbl
\begin{thebibliography}{10}
\expandafter\ifx\csname url\endcsname\relax
  \def\url#1{\texttt{#1}}\fi
\expandafter\ifx\csname urlprefix\endcsname\relax\def\urlprefix{URL }\fi
\expandafter\ifx\csname href\endcsname\relax
  \def\href#1#2{#2} \def\path#1{#1}\fi

\bibitem{Kaplan:2008}
J.~Kaplan, W.~Forrest, N.~Kindler, Revolutionizing data center energy
  efficiency, Tech. Rep. July, McKinsey \& Company (2008).

\bibitem{thermal:koomey2011}
J.~Koomey, Growth in data center electricity use 2005 to 2010, Tech. rep.,
  Analytics Press, Oakland, CA (2011).

\bibitem{Venkatraman:2012}
{Archana Venkatraman. ComputerWeekly.com}, Global census shows datacentre power
  demand grew 63\% in 2012,
  http://www.computerweekly.com/news/2240164589/Datacentre-power-demand-grew-63-in-2012-Global-datacentre-census
  (October 2012).

\bibitem{Breen:ITHERM:2010}
T.~Breen, et~al., From chip to cooling tower data center modeling: Part {I}
  influence of server inlet temperature and temperature rise across cabinet,
  in: ITherm, 2010, pp. 1--10.

\bibitem{uptimeInstitute:2013}
J.~K. Matt~Stansberry, Uptime institute 2013 data center industry survey, Tech.
  rep., Uptime Institute (2013).

\bibitem{El-Sayed:2012:TMD}
N.~El-Sayed, et~al., Temperature management in data centers: why some (might)
  like it hot, in: SIGMETRICS, 2012, pp. 163--174.

\bibitem{Brandon:GoingGreen:07}
J.~Brandon, Going green in the data center: Practical steps for your {SME} to
  become more environmentally friendly, Processor~(29) (2007).

\bibitem{Miller:Google:2012}
R.~Miller, Too hot for humans, but google servers keep humming,
  http://www.datacenterknowledge.com/archives/2012/03/23/too-hot-for-humans-but-google-servers-keep-humming/
  (March 2012).

\bibitem{ASHRAE:2011}
{ASHRAE Technical Commitee (TC) 9.9}, {2011 Thermal Guidelines for Data
  Processing Environments}, Tech. rep., American Society of Heating,
  Refrigerating and Air-Conditioning Engineers, Inc. (2011).

\bibitem{Ryan:LNCS:1998}
C.~Ryan, J.~Collins, M.~Neill, Grammatical evolution: Evolving programs for an
  arbitrary language, in: Genetic Programming, Vol. 1391 of Lecture Notes in
  Computer Science, Springer Berlin Heidelberg, 1998, pp. 83--96.

\bibitem{thermal:AtienzaReliable2008}
D.~Atienza, et~al., {Reliability-aware design for nanometer-scale devices}, in:
  Proceedings of the 2008 Asia and South Pacific Design Automation Conference,
  IEEE Computer Society Press, 2008, pp. 549--554.

\bibitem{Zapater:TPDS:2014}
M.~Zapater, et~al., Leakage-aware cooling management for improving server
  energy efficiency, IEEE Transactions on Parallel and Distributed Systems
  (TPDS) (2014).

\bibitem{Miller:2007}
R.~Miller, Data center cooling set points debated,
  http://www.datacenterknowledge.com/archives/2007/09/24/
  data-center-cooling-set-points-debated/ (September 2007).

\bibitem{Marshall:WP:2011}
P.~B. Liz~Marshall, Using {CFD} for data center design and analysis, Tech.
  rep., Applied Math Modeling White Paper (2011).

\bibitem{Abbasi:HPDC:2010}
Z.~Abbasi, G.~Varsamopoulos, S.~K.~S. Gupta, Thermal aware server provisioning
  and workload distribution for internet data centers, in: HPDC, ACM, New York,
  NY, USA, 2010, pp. 130--141.
\newblock \href {https://doi.org/10.1145/1851476.1851493}
  {\path{doi:10.1145/1851476.1851493}}.

\bibitem{Chen:2012:RTSS}
J.~Chen, et~al., A high-fidelity temperature distribution forecasting system
  for data centers, in: Proceedings of the 2012 IEEE 33rd Real-Time Systems
  Symposium, RTSS '12, IEEE Computer Society, Washington, DC, USA, 2012, pp.
  215--224.

\bibitem{Abbasi:Springer:2013}
Z.~Abbasi, M.~Jonas, A.~Banerjee, S.~Gupta, G.~Varsamopoulos, Evolutionary
  green computing solutions for distributed cyber physical systems, in:
  Evolutionary Based Solutions for Green Computing, Springer Berlin Heidelberg,
  2013, pp. 1--28.

\bibitem{Pagan:DCIS:2013}
J.~Pag\'{a}n, M.~Zapater, O.~Cubo, P.~Arroba, V.~Mart\'{\i}n, J.~M. Moya, A
  {Cyber-Physical} approach to combined {HW}-{SW} monitoring for improving
  energy efficiency in data centers, in: Conference on Design of Circuits and
  Integrated Systems, 2013, pp. 140--145.

\bibitem{Varsamopoulos:Springer:2009}
G.~Varsamopoulos, A.~Banerjee, S.~Gupta, Energy efficiency of thermal-aware job
  scheduling algorithms under various cooling models, in: Contemporary
  Computing, Vol.~40 of Communications in Computer and Information Science,
  2009, pp. 568--580.

\bibitem{Moore:ICAC:2006}
J.~Moore, J.~Chase, P.~Ranganathan, Weatherman: Automated, online and
  predictive thermal mapping and management for data centers, in: IEEE
  International Conference on Autonomic Computing, ICAC'06, 2006, pp. 155--164.
\newblock \href {https://doi.org/10.1109/ICAC.2006.1662394}
  {\path{doi:10.1109/ICAC.2006.1662394}}.

\bibitem{Silva:2013:ICMLA}
A.~M.~D. Silva, F.~Noorian, R.~I.~A. Davis, P.~H.~W. Leong, A hybrid feature
  selection and generation algorithm for electricity load prediction using
  grammatical evolution, in: Proceedings of the 2013 12th International
  Conference on Machine Learning and Applications, ICMLA '13, IEEE Computer
  Society, Washington, DC, USA, 2013, pp. 211--217.

\bibitem{Patterson:ITHERM:2008}
M.~Patterson, The effect of data center temperature on energy efficiency, in:
  Thermal and Thermomechanical Phenomena in Electronic Systems, ITHERM'08,
  2008, pp. 1167 --1174.

\bibitem{Heath:ASPLOS:2006}
T.~Heath, A.~P. Centeno, P.~George, L.~Ramos, Y.~Jaluria, R.~Bianchini, Mercury
  and freon: Temperature emulation and management for server systems, in:
  ASPLOS, New York, NY, USA, 2006, pp. 106--116.

\bibitem{Coskun:TCAD:2009}
A.~Coskun, T.~Rosing, K.~Gross, Utilizing predictors for efficient thermal
  management in multiprocessor socs, TCAD 28~(10) (2009) 1503 --1516.

\bibitem{Vladislavleva:TEVC:2009}
E.~Vladislavleva, G.~Smits, D.~den Hertog, Order of nonlinearity as a
  complexity measure for models generated by symbolic regression via pareto
  genetic programming, IEEE Transactions on Evolutionary Computation 13~(2)
  (2009) 333--349.
\newblock \href {https://doi.org/10.1109/TEVC.2008.926486}
  {\path{doi:10.1109/TEVC.2008.926486}}.

\bibitem{ONeill:TEVC:2001}
M.~O'Neill, C.~Ryan, Grammatical evolution, IEEE Transactions on Evolutionary
  Computation 5~(4) (2001) 349--358.

\bibitem{Back:TEVC:1997}
T.~Back, U.~Hammel, H.-P. Schwefel, Evolutionary computation: comments on the
  history and current state, IEEE Transactions on Evolutionary Computation
  1~(1) (1997) 3--17.
\newblock \href {https://doi.org/10.1109/4235.585888}
  {\path{doi:10.1109/4235.585888}}.

\bibitem{Kureichick:1996}
K.~Melikhov, V.~M. Kureichick, A.~N. Melikhov, V.~V. Miagkikh, O.~V. Savelev,
  A.~P. Topchy, {Some New Features In Genetic Solution Of The Traveling
  Salesman Problem.}, in: Adaptive Computing in Engineering Design and Control
  (ACEDC), 1996.

\bibitem{Rocha:Springer:1999}
M.~Rocha, J.~Neves, Preventing premature convergence to local optima in genetic
  algorithms via random offspring generation, in: International Conference on
  Industrial and Engineering Applications of Artificial Intelligence and Expert
  Systems, IEA/AIE'99, Secaucus, NJ, USA, 1999, pp. 127--136.

\bibitem{SPECCPU}
{SPEC CPU Subcommittee and John L. Henning}, {SPEC CPU 2006} benchmark
  descriptions, \url{http://www.spec.org/cpu2006/}.

\bibitem{Phansalkar:2007:SIGARCH}
A.~Phansalkar, A.~Joshi, L.~K. John, Subsetting the spec cpu2006 benchmark
  suite, SIGARCH Computer Architecture News 35~(1) (2007) 69--76.

\bibitem{Ryan:LBPGP:1998}
C.~Ryan, M.~O'Neill, Grammatical evolution: A steady state approach., in: In
  Late Breaking Papers, Genetic Programming, 1998, pp. 180--185.

\bibitem{Hemberg:GPEM:2013}
E.~Hemberg, L.~Ho, M.~O’Neill, H.~Claussen, A comparison of grammatical
  genetic programming grammars for controlling femtocell network coverage,
  Genetic Programming and Evolvable Machines 14~(1) (2013) 65--93.
\newblock \href {https://doi.org/10.1007/s10710-012-9171-8}
  {\path{doi:10.1007/s10710-012-9171-8}}.

\bibitem{Arroba:JGC:2014}
P.~Arroba, J.~L. Risco-Martin, M.~Zapater, J.~M. Moya, J.~L. Ayala, Enhancing
  regression models for complex systems using evolutionary techniques for
  feature engineering, Journal of Grid Computing (2014).

\end{thebibliography}
